\documentclass[%
 reprint,
superscriptaddress,
 amsmath,amssymb,
 aps,
]{revtex4-2}

\usepackage{xcolor}%
\usepackage{hyperref}
\usepackage{graphicx}
\usepackage{dcolumn}
\usepackage{bm}
\usepackage{ulem}
\usepackage{siunitx}

\begin{document}

\preprint{APS/123-QED}

\title{RheOPT: Few-Shot Non-Newtonian Flow Prediction \\
via Newtonian Pretraining of RheOFormer}

\author{Maedeh Saberi}
\affiliation{\mbox{Department of Mechanical and Industrial Engineering, Northeastern University, Boston, Massachusetts 02115, USA}}
\author{Amir Barati Farimani}
\affiliation{Department of Mechanical Engineering, Carnegie Mellon University, Pittsburgh, Pennsylvania 15213, USA}

\author{Safa Jamali}%
 \email{s.jamali@northeastern.edu}
\affiliation{\mbox{Department of Mechanical and Industrial Engineering, Northeastern University, Boston, Massachusetts 02115, USA}}


\begin{abstract}

Despite the advances in numerical simulations, non-Newtonian flow simulations have remained computationally expensive because the velocity and stress fields must be advanced together through rate- and history-dependent constitutive dynamics.
Neural-operator surrogates can reduce inference cost after training, but their training data often require the same expensive simulations they are intended to replace.
This makes data efficient learning especially important in regimes where only a small number of non-Newtonian simulations can be afforded.
Here, we introduce the Rheological Operator Pretrained Transformer (RheOPT), a neural-operator framework that pretrains a transformer-based operator on inexpensive Newtonian simulations and then fine-tunes it on a small number of non-Newtonian samples.
Rather than transferring the Newtonian constitutive response itself, RheOPT reuses geometry-conditioned transport structure that remains useful after the stress model changes.
We evaluate RheOPT in two geometry-matched settings: a $4{:}1$ contraction channel for Giesekus, Oldroyd-B, and Bingham fluids, and the oscillatory wake behind a triangular obstacle for Oldroyd-B flow.
In the contraction geometry, Newtonian pretraining substantially improves few-shot prediction, and a model fine-tuned on four target samples can match or exceed a from-scratch model trained on four times as many.
This benefit is reliable under reversible instance normalization (RevIN) but dataset dependent under Global normalization, a difference we trace to the dynamic range of each fluid's velocity and stress fields.
In the oscillatory wake, relative $L_2$ error provides a useful first comparison but cannot by itself verify that vortex shedding is recovered.
We therefore complement it with phase-insensitive diagnostics based on shedding frequency and wake fluctuation amplitude.
These diagnostics show that pretraining recovers the shedding frequencies, spatial structures, and the wake fluctuation amplitude that are suppressed by the model trained from scratch in the few-shot regime.
Overall, RheOPT establishes Newtonian simulation as a physically grounded pretraining source for data efficient non-Newtonian operator learning.

\end{abstract}


\maketitle


\section{Introduction}
\label{sec:introduction}

Predicting non-Newtonian flow is essential in polymer manufacturing~\cite{banerjee2023role}, biological systems~\cite{levental2007soft}, soft materials~\cite{mezzenga2005understanding}, and additive manufacturing~\cite{das2021importance, truby2016printing}, but it remains among the most computationally demanding problems in continuum mechanics because the stress field is itself a dynamic unknown~\cite{keunings1990progress}. 
Unlike Newtonian fluids, whose stress response is fixed by a viscosity, complex fluids couple the momentum balance to a rate- and history-dependent constitutive model. 
The internal stress must therefore be evolved together with the velocity field. 
Resolving this coupling with finite-element or finite-volume solvers is expensive, and the cost must be paid again whenever the geometry, boundary condition, or constitutive parameter changes, since each scenario requires a full re-computation~\cite{alves2021numerical, walters2003distinctive}.

This cost has motivated the development of machine-learning surrogates, and neural operators in particular~\cite{lu2021learning, li2020fourier, kovachki2023neural, azizzadenesheli2024neural}, which learn solution maps between function spaces and can therefore predict flow fields for new conditions at inference cost rather than solver cost.
Our prior work, RheOFormer~\cite{saberi2025rheoformer}, established that a transformer-based neural operator can reproduce spatio-temporal stress and velocity fields in viscoelastic and viscoplastic flows with complex geometries. 
Reaching this accuracy, however, comes with a large data requirement, as neural operators are data-hungry and typically require many training simulations spanning the range of conditions the model must later predict.

This creates a practical bottleneck.
If many expensive non-Newtonian simulations are needed before the surrogate can be trained, the main computational burden is shifted from inference to data generation rather than removed.
A learned model is therefore most valuable in the regime where it is also hardest to train: the data-scarce setting in which only a handful of non-Newtonian simulations can be afforded.

Newtonian simulations offer a physically motivated way to address this limitation.
They are typically easier and more robust to generate than non-Newtonian simulations since they avoid the additional constitutive evolution and numerical stiffness associated with complex-fluid stresses. 
Meanwhile, they share important structure with non-Newtonian flows as the geometry, boundary conditions, and momentum balance organize the velocity gradients and transport pathways in both cases. 
These structures provide the kinematic scaffold on which viscoelastic or viscoplastic stresses evolve. 
This motivates the central question of this work: can a neural operator pretrained on Newtonian simulations be fine-tuned, using only a few non-Newtonian samples, into an accurate surrogate for complex-fluid flow?

This transfer is not guaranteed. 
Pretraining and fine-tuning are separated by a change in the constitutive law itself.
That change is the defining difference between Newtonian and non-Newtonian flow, and it is not obvious that a representation learned under a Newtonian stress response survives the transition and remains useful for viscoelastic or viscoplastic prediction. 
Our hypothesis is that Newtonian pretraining transfers not the Newtonian constitutive response itself, but representations of momentum transport, velocity gradients, and boundary-driven flow organization that remain relevant after the stress model changes.
Prior studies for partial differential equation (PDE) surrogates have transferred representations across coefficients, fidelities, and Newtonian equation families~\cite{mccabe2023multiple, subramanian2023towards, zolfaghari2026hierarchical, goswami2022deep}, while data-driven models of non-Newtonian flow have generally been trained on data from the target constitutive model~\cite{thakur2024viscoelasticnet, mahmoudabadbozchelou2022nn, mangal2025learning}. 
To the best of our knowledge, Newtonian to non-Newtonian transfer across a constitutive law boundary has not been systematically examined, and it is an open question that RheOPT is designed to answer.

Different factors may determine when this transfer can be relied upon. 
One thing that should be considered is that non-Newtonian datasets can span wide ranges of velocity and stress magnitudes across flow regimes. 
The normalization used before training can therefore affect whether the pretrained representation remains useful during fine-tuning or is obscured by scale mismatch between the Newtonian source data and the non-Newtonian target data~\cite{kim2021reversible}.
Second, many complex-fluid flows are unsteady, so pointwise error alone may not determine whether the relevant dynamics have been recovered.
In oscillatory regimes such as vortex shedding, a prediction with the correct shedding structure can be penalized by a phase shift, while a nearly steady prediction can appear deceptively accurate if it remains close to the time-averaged field.
Such cases require diagnostics that test the temporal content of the flow, not only the total field error~\cite{hu2025generative, um2020solver}.
These issues are not separate from transfer. 
They determine whether transferred structure is both visible in the trained model and physically meaningful in the predicted flow.

We introduce RheOPT, a Newtonian-pretraining framework for few-shot non-Newtonian neural-operator prediction.
We evaluate the framework in two canonical geometries:
a $4{:}1$ contraction channel and a flow past a triangular obstacle.
The $4{:}1$ contraction channel provides a controlled test of physical constitutive transfer across Giesekus, Oldroyd-B, and Bingham fluids, while the triangular obstacle wake tests whether the pretrained representations remain useful for recovering complex vortex-shedding dynamics.

The main contribution of this work is a controlled test of Newtonian to non-Newtonian transfer for neural-operator prediction.
In the $4{:}1$ contraction geometry, we show that geometry-matched Newtonian pretraining improves few-shot prediction across Giesekus, Oldroyd-B, and Bingham fluids, especially in the strict two snapshot input setting.
We then show that the reliability of this transfer depends on field normalization: reversible instance normalization (RevIN)~\cite{kim2021reversible} gives consistent gains across datasets with wide field-magnitude variation, whereas Global normalization is more dataset dependent.
Finally, we extend the analysis from quasi-steady contraction flow to the oscillatory wake flow past a triangular obstacle and use phase-insensitive diagnostics to show that pretraining helps recover vortex-shedding dynamics that are suppressed in the from-scratch model.

\section{Related Work}
\label{sec:related_work}

Data-driven modeling of complex fluids has developed rapidly, driven by the difficulty of characterizing rate- and history-dependent rheology from limited measurements.
A substantial line of work embeds constitutive physics directly into the learning objective.
Rheology-informed neural networks and related extensions recover velocity and stress fields, and can identify constitutive parameters, from sparse rheometric or flow data~\cite{mahmoudabadbozchelou2021rheology, zolfaghari2026non}. 
Other frameworks such as ViscoelasticNet use learned models to select among candidate viscoelastic constitutive laws and reconstruct the stress field from an observed velocity field~\cite{thakur2024viscoelasticnet}.
Related approaches learn constitutive response from deformation history~\cite{mahmoudabadbozchelou2022digital}, and data-driven surrogates have also been used for more complex stress prediction tasks~\cite{balasubramanian2025prediction}.
However, across these methods the model is usually trained for a specific fluid, constitutive family, or flow configuration, and a change in constitutive model or geometry requires a new target dataset and retraining.

Operator learning addresses a different part of this challenge. 
Rather than approximating one solution, neural operators learn the mapping between input and output functions. 
They can therefore predict solutions for new conditions without retraining, provided those conditions lie within the learned distribution~\cite{lu2021learning, li2020fourier, kovachki2023neural}. 
This idea has been used to learn families of rheological constitutive models and to generalize across parameters within a given family~\cite{mangal2025learning}. 
Attention-based and geometry-aware operator architectures further improve expressivity and support irregular meshes in complex geometries~\cite{li2022transformer, hao2023gnot, wu2024transolver, li2023geometry}. 
RheOFormer extended this direction to rheology by learning full spatio-temporal stress and velocity evolution in viscoelastic and viscoplastic flows in irregular geometries~\cite{saberi2025rheoformer}. 
These models reduce inference cost after training and can generalize within the distribution represented by their training data.
However, their training sets are still drawn from the target physics, so data generation remains a central cost when high-fidelity non-Newtonian simulations are scarce.

Pretraining has emerged as a central strategy for large language models and has recently become an active direction in scientific machine learning for reducing downstream data requirements.
Prior work has transferred neural-operator representations across coefficients, boundary conditions, and equation families~\cite{subramanian2023towards, goswami2022deep}. 
A systematic study of pretraining strategies for neural operators found that transfer learning is consistently among the most effective approaches, and it is most beneficial when downstream data are scarce~\cite{zhou2024strategies}. 
Large multi-physics models extend the idea further by pretraining on heterogeneous PDE systems using shared embeddings and normalization strategies that accommodate variation across physics domains~\cite{mccabe2023multiple, mccabe2024multiple, hao2024dpot, herde2024poseidon}. 
Recent work has also explored pretraining on cheaper lower-dimensional data before transferring to more expensive target problems~\cite{hemmasian2024pretraining}. 
RheOPT follows this data-efficiency motivation, but uses a different and physically grounded source of cheaper data.

Transfer learning has also been studied in physics-informed neural networks. 
In this setting, a network trained on one problem can initialize training on another through lightweight adaptation~\cite{tang2022transfer}. 
These methods can accelerate optimization and convergence when source and target tasks are related, but they are usually confined to instances of a single physical problem, such as changes of parameters~\cite{gao2022svd}, boundary conditions~\cite{goswami2020transfer}, or fidelity~\cite{chakraborty2021mfpidnn}.
These methods can reduce optimization cost, but they generally still require target-specific re-optimization.

Across these directions, a common gap remains. 
Existing non-Newtonian learning methods generally train on data from the same target fluid, constitutive family, or flow configuration.
Neural operators can be geometrically flexible and generalizable across input families, but their training data are still drawn from the target physics. 
PDE pretraining studies demonstrate transfer across several forms of distribution shift, but they have not systematically tested whether a representation learned from Newtonian flow can be transferred and remain useful for viscoelastic and viscoplastic prediction.
This is exactly the gap that RheOPT is designed to address by using Newtonian simulations as a physically grounded pretraining source for few-shot non-Newtonian operator learning.

\section{Methods}
\label{sec:methods}

\subsection{Problem setup and transfer design}
\label{sec:problem_transfer}

We consider the data-efficient prediction of incompressible, two-dimensional non-Newtonian flows on irregular domains.
Each domain is represented by an unstructured mesh with node coordinates $(x,y)$.
At each node, the flow state contains the velocity components $(v_x,v_y)$ and the independent components of the symmetric stress tensor $(\sigma_{xx},\sigma_{xy},\sigma_{yy})$.
The model receives a short input sequence of these five physical fields, together with the node coordinates, and predicts the remaining spatio-temporal trajectory.

The input window and prediction horizon depend on the geometry.
For the contraction experiments, the model receives two snapshots over $t\in[0,0.2]~\mathrm{s}$, sampled at $\Delta t=0.2~\mathrm{s}$, and predicts the subsequent evolution up to $t=5~\mathrm{s}$.
For the triangular-obstacle experiments, the model receives ten snapshots over $t\in[0,1.8]~\mathrm{s}$, also sampled at $\Delta t=0.2~\mathrm{s}$, and predicts the remaining trajectory up to $t=10~\mathrm{s}$.
These two settings provide complementary tests: the contraction geometry evaluates constitutive transfer in a short observation regime, while the triangular-obstacle geometry evaluates transfer in an unsteady wake with vortex shedding.

\begin{figure*}[t]
    \centering
    \includegraphics[width=\textwidth]{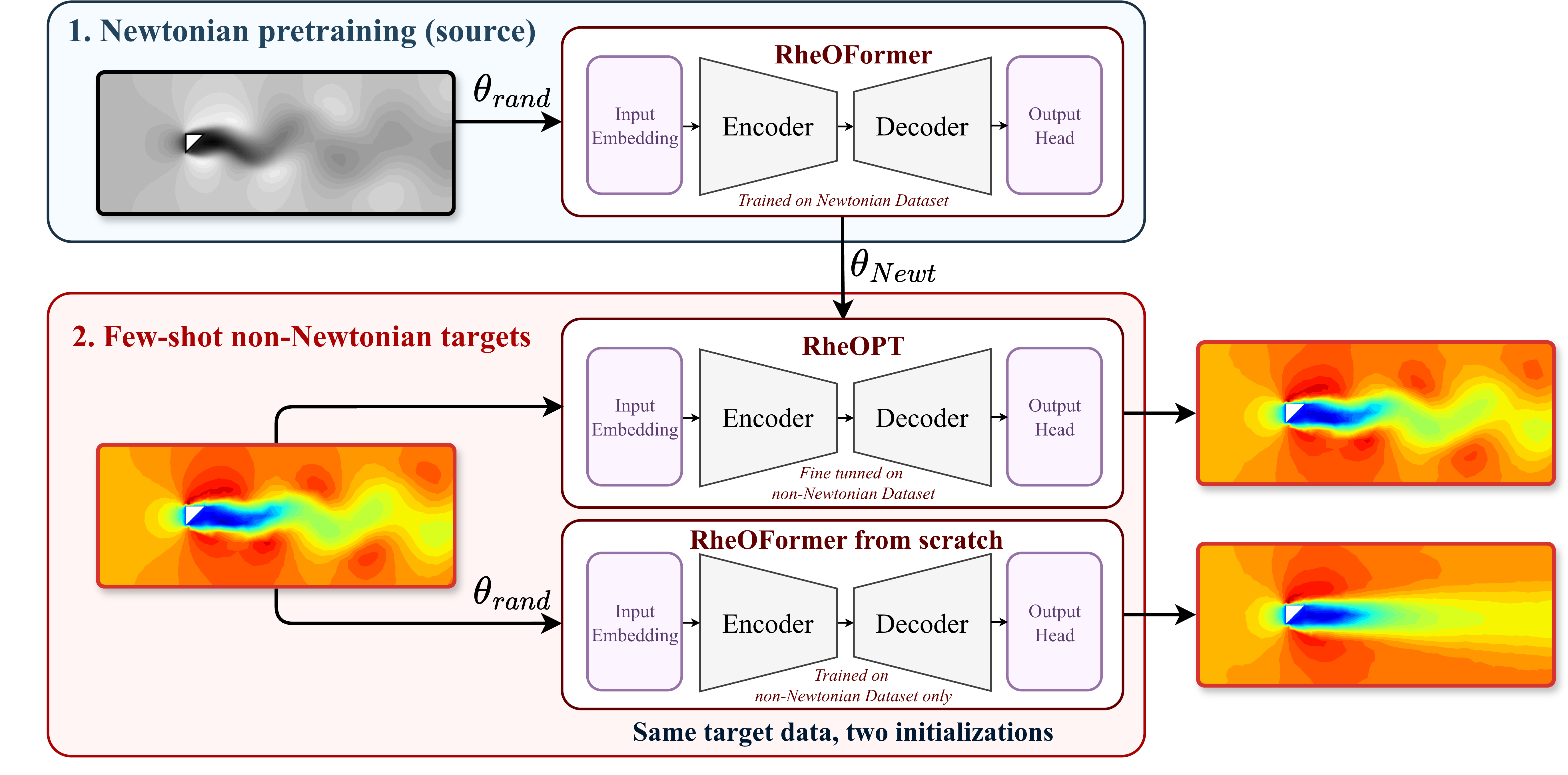}
    \caption{\textbf{Overview of RheOPT.} (1) RheOFormer is trained from a random initialization, $\theta_{\mathrm{rand}}$, on Newtonian simulations of the target geometry, yielding the pretrained weights $\theta_{\mathrm{Newt}}$. (2) The same few-shot non-Newtonian target data are used to fine-tune the pretrained model (RheOPT, initialized with $\theta_{\mathrm{Newt}}$) and to train an otherwise identical RheOFormer from scratch (initialized with $\theta_{\mathrm{rand}}$). The fields illustrate the triangular-obstacle wake: a Newtonian snapshot used for pretraining (gray), an Oldroyd-B target snapshot (left), and the fine-tuned and from-scratch predictions (right).}
    \label{fig:schematic}
\end{figure*}

RheOPT is formulated as a geometry-matched transfer-learning problem.
For each geometry, the source task is Newtonian flow prediction and the target task is non-Newtonian flow prediction in the same domain.
The non-Newtonian targets are Giesekus and Oldroyd-B fluids for viscoelastic flow and a Bingham fluid for viscoplastic flow.
The source and target tasks therefore share the geometry, boundary-driven transport structure, and momentum balance, while differing in the constitutive relation that determines the stress response.
This design tests whether geometry and transport constrained structures, such as recirculation zones, stress concentrations near walls and corners, and wake formation, can be reused after the constitutive closure changes.
Figure~\ref{fig:schematic} summarizes this workflow. RheOFormer is first trained from a random initialization, $\theta_{\mathrm{rand}}$, on Newtonian simulations of a given geometry, which yields the pretrained weights $\theta_{\mathrm{Newt}}$. The same few non-Newtonian target samples are then used in two ways: RheOPT is initialized with $\theta_{\mathrm{Newt}}$ and fine-tuned on the target data, whereas the baseline trains the same architecture from $\theta_{\mathrm{rand}}$ on the target data only.

All fluids considered here satisfy the incompressibility constraint and momentum balance
\begin{equation}
\nabla \cdot \mathbf{u} = 0, \qquad
\rho\!\left(\frac{\partial \mathbf{u}}{\partial t}
+ \mathbf{u}\cdot\nabla\mathbf{u}\right)
= \nabla \cdot \boldsymbol{\sigma},
\label{eq:momentum}
\end{equation}
where $\mathbf{u}=(v_x,v_y)$ and $\boldsymbol{\sigma}$ is the total Cauchy stress tensor.
The stress can be decomposed as $\boldsymbol{\sigma}=-p\mathbf{I}+\boldsymbol{\tau}$,
where $p$ is the pressure and $\boldsymbol{\tau}$ is the non-hydrostatic stress contribution.
For a Newtonian fluid, $\boldsymbol{\tau}=2\eta\mathbf{D}$, with $\eta$ the constant viscosity and $\mathbf{D}=(\nabla\mathbf{u}+\nabla\mathbf{u}^{T})/2$ the rate-of-deformation tensor.
Equation~\eqref{eq:momentum} is common to every fluid considered here; the fluid models differ in the constitutive relation governing
$\boldsymbol{\tau}$ and, consequently, the resulting total stress $\boldsymbol{\sigma}$.
For the non-Newtonian targets, $\boldsymbol{\tau}$ is governed by the corresponding Giesekus, Oldroyd-B, or Bingham constitutive relation.
Thus, the transfer studied here is not across unrelated governing equations, but across constitutive closures within the same momentum-conservation framework.
The full constitutive relations are provided in the Supplementary Information (S1).

The geometry-matched design isolates constitutive transfer from geometric transfer.
The pretrained model is not asked to adapt to a new domain during fine-tuning; instead, it must adapt representations learned from Newtonian flow to the target viscoelastic or viscoplastic stress dynamics.
Table~\ref{tab:experimental_design} summarizes the source and target configurations used to test this transfer.
For each geometry, the source dataset consists of $200$ Newtonian simulations spanning inlet velocities corresponding to $Re\in[0.4,80]$ for the contraction and $Re\in[10,100]$ for the triangular obstacle.
The Newtonian and non-Newtonian simulations share the same domain but are discretized on different unstructured meshes. Because the encoder and decoder operate directly on the node coordinates, no interpolation between these meshes is required.
The downstream non-Newtonian models are fine-tuned using small target-data budgets and compared with from-scratch models trained on the same target samples.

The contraction geometry is the primary test case and is evaluated for all three target fluids.
In the Giesekus and Oldroyd-B datasets, the inlet velocity is varied, causing the Reynolds and Weissenberg numbers to change together.
In the Bingham dataset, the inlet velocity and yield stress are varied independently, so the data span both Reynolds and Bingham numbers.
For the contraction datasets, these dimensionless numbers are defined with the downstream scales, $Re=\rho U L/\eta_0$, $Wi=\lambda U/L$, and $Bi=\tau_y L/(\mu_p U)$, where $U=4U_{in}$ is the mean velocity in the downstream channel, $U_{in}$ is the mean inlet velocity, $L=0.01~\mathrm{m}$ is the half-width of the downstream channel, $\eta_0$ is the zero-shear viscosity (or plastic viscosity $\mu_p$ for the Bingham fluid), $\lambda$ is the relaxation time, $\tau_y$ is the yield stress, and $\mu_p$ is the plastic viscosity.
The triangular-obstacle geometry is evaluated for Oldroyd-B flow, with $Re\in[10,100]$ and $Wi=0.01\,Re$, and provides a more demanding dynamical case involving wake formation and periodic vortex shedding.
Exact parameter ranges, sampled values, mesh statistics, dataset partitions, and numerical-solver settings are provided in the Supplementary Information (S2).

\begin{table*}[t]
\centering
\caption{
Summary of the RheOPT experimental design.}
\label{tab:experimental_design}
\begin{tabular}{l|lcccc}
\hline
{Fluid model}
&
{Geometry}
&
{Varied \newline target \newline controls}
&
{Fine-tuning samples}
&
{Input window}
&
{Prediction horizon}
\\
\hline

Giesekus
&
$4{:}1$ contraction
&
Inlet velocity
&
$4,8,12,16$
&
$t\in[0,0.2]$ s
&
$t\in[0.4,5]$ s
\\

Oldroyd-B
&
$4{:}1$ contraction
&
Inlet velocity
&
$4,8,12,16$
&
$t\in[0,0.2]$ s
&
$t\in[0.4,5]$ s
\\

Bingham
&
$4{:}1$ contraction
&
Inlet velocity \& yield stress
&
$4,8,12,16$
&
$t\in[0,0.2]$ s
&
$t\in[0.4,5]$ s
\\

\hline
Oldroyd-B
&
Triangular obstacle
&
Inlet velocity
&
$4, 8, 12, 16, 24, 32$
&
$t\in[0,1.8]$ s
&
$t\in[2,10]$ s
\\

\hline
\end{tabular}
\end{table*}

\subsection{RheOPT architecture, inputs, and training}
\label{sec:model_training}

Realizing this transfer design requires an architecture that can operate on irregular meshes and propagate dynamics over long horizons. 
RheOPT builds upon the RheOFormer architecture~\cite{saberi2025rheoformer, li2022transformer} as its neural-operator backbone.
RheOFormer is a transformer-based encoder--decoder architecture designed for complex-fluid flow prediction on irregular meshes.
The encoder applies self-attention to capture spatial dependencies across all mesh nodes, producing a discretization-invariant latent representation of the observed flow state.
The decoder uses cross-attention between this latent representation and query-point coordinates to predict the output fields at arbitrary locations, decoupling the output discretization from the input mesh.
Temporal evolution is performed in latent space through a recurrent propagator, $z_{t+1}=z_t+\mathcal{N}(z_t)$,
which advances the latent state without repeatedly re-encoding physical fields at every timestep.
This allows the model to predict long trajectories on the contraction and triangular-obstacle meshes without remeshing, interpolation, or geometry-specific architectural changes.
Architectural hyperparameters follow RheOFormer~\cite{saberi2025rheoformer} and are provided in the Supplementary Information.

At each mesh node, the input feature vector consists of the five observed physical fields and the node coordinates,
\begin{equation}
\mathbf{x}_{\mathrm{in}}
=
\left[
v_x,\ v_y,\ \sigma_{xx},\ \sigma_{xy},\ \sigma_{yy},\ x,\ y
\right].
\label{eq:input}
\end{equation}
The coordinates identify the node locations on the irregular mesh, while the physical channels provide the observed initial trajectory.
Reynolds, Weissenberg, and Bingham numbers are not appended as additional node features.

We compare two input-scaling formulations: \textit{Global normalization} and \textit{reversible instance normalization (RevIN)}~\cite{kim2021reversible}.
Global normalization uses fixed per-channel statistics computed from the training dataset.
Each physical channel is standardized using its training-set mean and standard deviation, and the same statistics are used to normalize the inputs and targets and to transform predictions back to physical space.
For the pretrained models, these statistics are computed from the Newtonian pretraining dataset and are retained during fine-tuning, whereas the from-scratch models use the statistics of their non-Newtonian training samples.

RevIN instead computes normalization statistics separately for each sample and physical channel from the observed input sequence.
For sample $b$ and channel $c$, let $\mu_{b,c}$ and $s_{b,c}$ denote the corresponding mean and standard deviation.
These statistics are computed over all input snapshots and mesh nodes of the sample.
The input fields are normalized using these sample-specific statistics and transformed with learnable channel-wise affine parameters $\gamma_c$ and $\beta_c$:
\begin{equation}
\widetilde{f}_{b,t,n,c}
=
\gamma_c
\frac{f_{b,t,n,c}-\mu_{b,c}}{s_{b,c}}
+\beta_c .
\label{eq:revin_forward}
\end{equation}
The affine parameters are optimized jointly with the RheOFormer weights during pretraining and fine-tuning.
These affine parameters act only on the model input, and the target trajectory is normalized with the same input statistics, $(f_{b,t,n,c}-\mu_{b,c})/s_{b,c}$, without the affine transform.
After the model predicts the normalized future trajectory, the transformation is inverted using the same statistics computed from the input sequence,

{
\begin{equation}
\widehat{f}_{b,t,n,c} = \widehat{\widetilde{f}}_{b,t,n,c}\, s_{b,c} + \mu_{b,c}.
\label{eq:revin_inverse}
\end{equation}}
Thus, RevIN removes the sample-specific offset and scale before the trajectory is processed by the neural operator and restores them after prediction. 
This formulation reduces sample-to-sample shifts in the magnitude and offset of the physical fields while retaining a reversible mapping to the original representation.
In contrast, Global normalization transforms every sample using a single set of fixed, per-channel statistics computed from the training dataset. 

Following prior neural-operator pretraining studies~\cite{zhou2024strategies}, we use transfer learning with the same field prediction objective during Newtonian pretraining and non-Newtonian fine-tuning, rather than introducing a separate self-supervised pretraining task.
The loss is a relative error averaged over samples and output channels,
\begin{equation}
\begin{gathered}
\mathcal{L} =
\frac{1}{B \cdot C} \sum_{b=1}^{B} \sum_{c=1}^{C}
\frac{\lVert \hat{\mathbf{f}}_{b,:,:,c} - \mathbf{f}_{b,:,:,c} \rVert_2}
     {\lVert \mathbf{f}_{b,:,:,c} \rVert_2 + \epsilon}, \\[2mm]
\text{where} \qquad
\lVert \mathbf{f}_{b,:,:,c} \rVert_2
= \sqrt{\textstyle\sum_{t=1}^{T} \sum_{n=1}^{N_{\mathrm{node}}} \mathbf{f}_{b,t,n,c}^2}.
\end{gathered}
\label{eq:loss}
\end{equation}
Here, $\widehat{\mathbf{f}}$ and $\mathbf{f}$ are the predicted and ground-truth trajectories, $B$ is the batch size, $C$ is the number of output channels, $T$ is the number of predicted frames, $N_{\mathrm{node}}$ is the number of mesh nodes, and $\epsilon$ is a small constant used for numerical stability.
The norm is computed jointly over the temporal and spatial dimensions for each channel.

For each downstream experiment, the fine-tuned model is initialized from the corresponding geometry-matched Newtonian-pretrained weights.
All model parameters are updated during fine-tuning; no encoder, decoder, or latent-propagator layers are frozen.
Full fine-tuning allows the complete representation to adapt to the target constitutive behavior rather than restricting adaptation to a predetermined subset of layers.
Each fine-tuned model is compared with an otherwise identical from-scratch model trained from random initialization.
The pretrained and from-scratch models use the same RheOFormer architecture, the same non-Newtonian fine-tuning samples, the same normalization formulation, optimizer, learning-rate schedule, batch size, number of optimization steps, and validation set.
Thus, initialization from the Newtonian-pretrained weights is the only intended difference between the pretrained and from-scratch models.

For each target fluid and data budget, the fine-tuning samples are disjoint from the validation and evaluation sets.
The same validation and evaluation sets are used for the corresponding pretrained and from-scratch models.
All models are trained using the AdamW optimizer with a maximum learning rate of $10^{-4}$ and a one-cycle learning-rate scheduler with a warm-up over the first $30\%$ of the optimization steps, an initial division factor of $10^{4}$, and a final division factor of $10^{6}$.
During fine-tuning, the batch size is set to $4$.
Each model is trained for $500$ epochs in the contraction experiments and $2000$ epochs in the triangular-obstacle experiments, so that the number of optimization steps is identical for the pretrained and from-scratch models at a given $N$.
Each condition is repeated using four independent training seeds, and results are reported as the mean and standard deviation across these runs.

\subsection{Evaluation metrics and dynamical diagnostics}
\label{sec:evaluation}

All reported errors are computed in physical space after reversing the normalization associated with either Global normalization or RevIN.
This ensures that the two input scaling formulations are compared using the same physical variables rather than formulation dependent normalized coordinates.
Unless otherwise stated, errors are accumulated over the full predicted time horizon, averaged over the fixed validation set, and reported as the mean and standard deviation across three independent training seeds.

The principal evaluation metric is the real-space relative $L_2$ error.
For a scalar field $f$, the error is computed over all predicted timesteps and mesh nodes as
\begin{equation}
\mathcal{E}_{L_2}(f)
=
\frac{
\left[
\sum_{t=1}^{T_{\mathrm{pred}}}
\sum_{n=1}^{N_{\mathrm{node}}}
\left(
\hat{f}_{t,n}-f_{t,n}
\right)^2
\right]^{1/2}
}{
\left[
\sum_{t=1}^{T_{\mathrm{pred}}}
\sum_{n=1}^{N_{\mathrm{node}}}
f_{t,n}^{2}
\right]^{1/2}
+\epsilon
}.
\label{eq:relative_l2_eval}
\end{equation}
Here, $T_{\mathrm{pred}}$ is the number of predicted timesteps, $N_{\mathrm{node}}$ is the number of mesh nodes, and $\epsilon$ is a small constant used for numerical stability.
The velocity error is obtained by evaluating Eq.~\eqref{eq:relative_l2_eval} jointly over $(v_x,v_y)$, and the stress error by evaluating it jointly over $(\sigma_{xx},\sigma_{xy},\sigma_{yy})$, so that the sums also run over the components of each group. These errors are computed for each validation sample and then averaged over the validation set.
This separation prevents the different numerical scales and physical behavior of velocity and stress from obscuring field-specific effects of pretraining.

To quantify the benefit of Newtonian pretraining relative to the from-scratch baseline, we define the percentage improvement as
\begin{equation}
\mathrm{Improvement}(\%) = 100 \frac{L_{\mathrm{scratch}} - L_{\mathrm{pretrained}}}{L_{\mathrm{scratch}}},
\label{eq:improvement}
\end{equation}
where $L_{\mathrm{scratch}}$ and $L_{\mathrm{pretrained}}$ are the corresponding real-space relative $L_2$ errors.
Positive values indicate that Newtonian pretraining reduces the prediction error relative to training from random initialization.

For the triangular-obstacle wake, relative $L_2$ error is complemented by phase-insensitive diagnostics that directly measure vortex shedding dynamics.
This is necessary because a prediction with the correct oscillatory structure can be penalized by phase or spatial offsets, while a smoother prediction can appear accurate if it remains close to the time-averaged wake.
We therefore evaluate the dominant shedding frequency and the wake-fluctuation amplitude.

The dominant shedding frequency is extracted from the temporal spectrum of the predicted wake signal and compared with the ground-truth frequency across validation samples with different Reynolds numbers.
For each wake node $n\in\mathcal{W}$, the temporal mean of the streamwise velocity is removed and its discrete Fourier transform is computed over the predicted horizon,
\begin{equation}
\begin{gathered}
\hat{v}_{x,n}(f_k)=\sum_{t=1}^{T_{\mathrm{pred}}}\left[v_x(\mathbf{x}_n,t)-\overline{v_x}(\mathbf{x}_n)\right]e^{-2\pi i f_k t\Delta t},\\
\text{and}
\qquad f_k=\frac{k}{T_{\mathrm{pred}}\,\Delta t},
\label{eq:wake_dft}
\end{gathered}
\end{equation}
where $\mathcal{W}$ is the set of $N_{\mathcal{W}}$ wake nodes with $x>1$ and $|y|<0.5$, and $x$ is measured from the channel inlet.
The wake power spectrum and the dominant shedding frequency are then

\begin{equation}
\begin{gathered}
P(f_k)=\frac{1}{N_{\mathcal{W}}\,T_{\mathrm{pred}}}\sum_{n\in\mathcal{W}}\left|\hat{v}_{x,n}(f_k)\right|^2,\\
\text{and}
\qquad
f_{\mathrm{peak}}=\underset{f_k,\;k\ge1}{\arg\max}\;P(f_k).
\label{eq:peak_frequency}
\end{gathered}
\end{equation}
With $T_{\mathrm{pred}}=41$ snapshots at $\Delta t=0.2~\mathrm{s}$, the frequency resolution is $1/(T_{\mathrm{pred}}\Delta t)\approx0.12~\mathrm{Hz}$, so $f_{\mathrm{peak}}$ at the lowest non-zero frequency indicates the absence of a resolved oscillation rather than a physical low-frequency shedding mode.
This diagnostic tests whether the model recovers the characteristic time scale of vortex shedding.

To measure the strength and spatial persistence of the oscillation, we also compute the root-mean-square fluctuation amplitude of the transverse velocity in the wake,
\begin{equation}
v_{y,\mathrm{rms}}'(x)
=
\left[
\frac{1}{T_{\mathrm{pred}}}
\sum_{t=1}^{T_{\mathrm{pred}}}
\left(
v_y(x,t)-\overline{v_y}(x)
\right)^2
\right]^{1/2},
\label{eq:rms_fluctuation}
\end{equation}
where $\overline{v_y}(x)$ is the temporal mean at each downstream location.
In practice, $v_{y,\mathrm{rms}}'$ is evaluated in streamwise bins of width $0.1$ by averaging the temporal variance of $v_y$ over all nodes in each bin before taking the square root.
Together, the shedding-frequency and RMS-fluctuation diagnostics evaluate whether the predicted wake contains the correct temporal scale and fluctuation strength, rather than only a low pointwise error.

\section{Results and Discussion}\label{results}

\begin{figure*}[t]
    \centering
    \includegraphics[width=\textwidth]{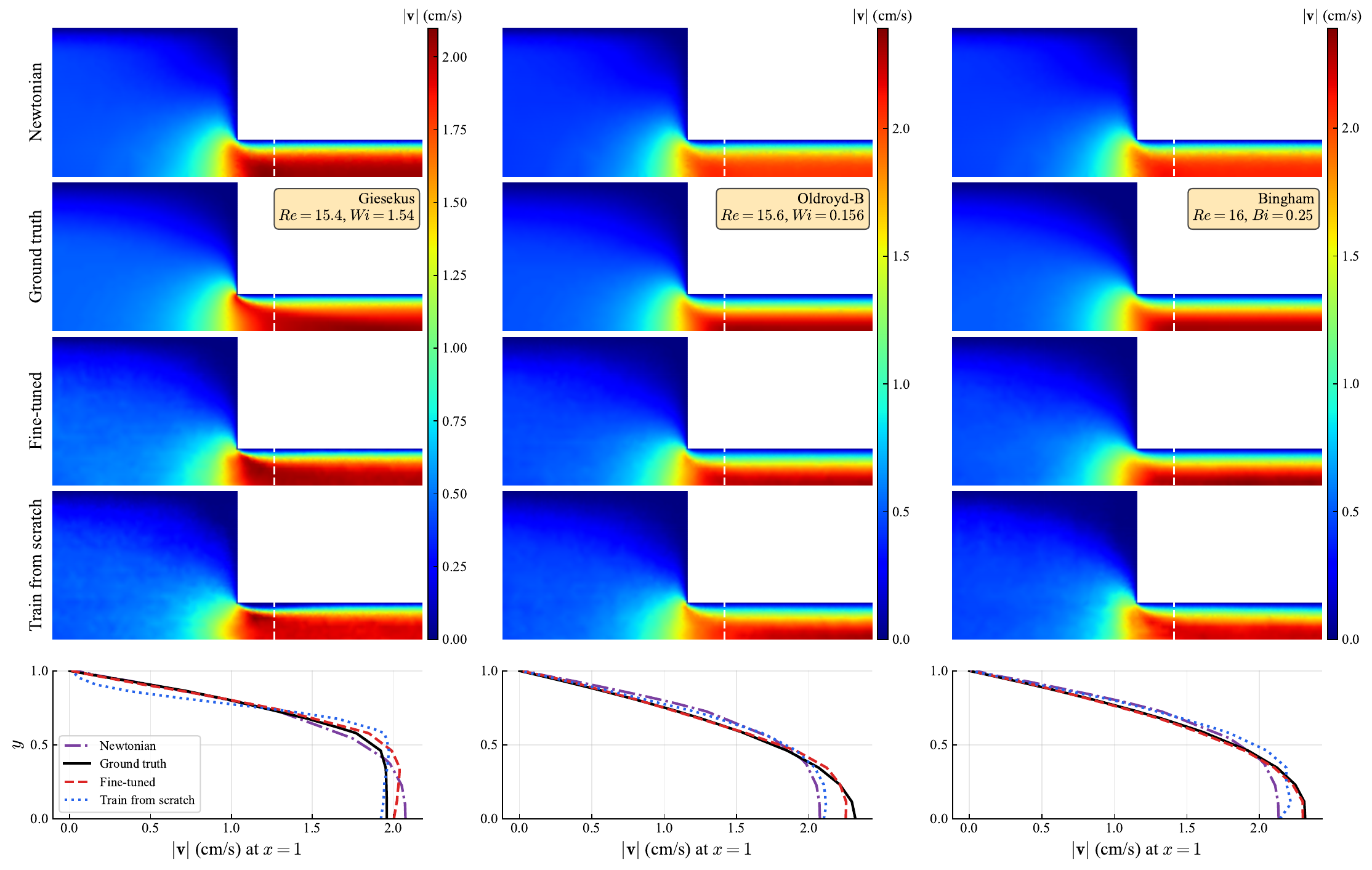}\caption{\textbf{Representative contraction-flow velocity fields.} Velocity magnitude, $|\mathbf{v}|$, at $t=5~\mathrm{s}$ for Giesekus ($Re=15.4$, $Wi=1.54$), Oldroyd-B ($Re=15.6$, $Wi=0.156$), and Bingham ($Re=16$, $Bi=0.25$) fluids in the $4{:}1$ contraction, predicted with $N=4$ fine-tuning samples under RevIN. Rows show the Newtonian pretraining simulation at the same $Re$, the ground truth, the fine-tuned (pretrained) prediction, and the from-scratch prediction; all fields of a column share one color scale. The bottom row shows the corresponding profiles across the downstream channel at $x=1$, marked by the dashed lines, one downstream half-width past the contraction plane. The corresponding error maps and the shear-stress fields are shown in Figs.~S2 and S3 of the Supplementary Information.}
    \label{fig:contraction_fields}
\end{figure*}

We organize the results around one central question: does Newtonian pretraining transfer reusable flow structure to few-shot non-Newtonian prediction, and under what conditions is that transfer visible?
We first use the geometry-matched $4{:}1$ contraction as a controlled setting, where the model has only a short observation window and any improvement over an identically trained from-scratch baseline can be attributed to information carried by the Newtonian initialization. 
We then show that this transfer is not independent of the input representation: Global normalization and RevIN reveal different levels of reliability because the target datasets differ in field-magnitude range. 
Finally, we test an oscillatory triangular-obstacle wake, where relative $L_2$ error must be complemented by phase-insensitive diagnostics to determine whether vortex-shedding dynamics are recovered. 
Throughout, errors are reported as real-space relative $L_2$ after reversing normalization, and every pretrained model is compared with a from-scratch model trained with the same architecture, target data, optimizer, schedule, and random-seed protocol. 
When reported, ``improvement'' denotes the percentage reduction in error obtained by Newtonian pretraining relative to the from-scratch baseline, as defined in Sec.~\ref{sec:evaluation}.

\subsection{Newtonian pretraining improves few-shot contraction-flow prediction}
\label{sec:results_contraction_transfer}

We begin with the $4{:}1$ contraction geometry as the most controlled test of constitutive transfer, where the Newtonian pretraining simulations and non-Newtonian fine-tuning simulations share the same domain.
This geometry-matched setting provides the cleanest test of the transfer hypothesis: the model is not asked to extrapolate to a new geometry, but to adapt representations learned from Newtonian momentum transport to flows governed by Giesekus, Oldroyd-B, or Bingham constitutive behavior. For each target fluid, the pretrained model is fully fine-tuned using only $N\in\{4,8,12,16\}$ non-Newtonian simulations and is compared with an identical model trained from scratch using the same target samples, architecture, optimizer, and training schedule.

The contraction experiments are deliberately data-limited in time.
The model observes only two snapshots before predicting the subsequent evolution.
This choice is important because a long input history can already contain substantial information for a from-scratch model to infer much of the time evolution directly from the observed trajectory. 
By restricting the input to two snapshots, we reduce the observed temporal history and force the prediction to rely more strongly on structure learned across the training distribution.
This setting therefore provides a stricter and more informative test of whether Newtonian pretraining transfers useful momentum-transport and geometry-conditioned information to non-Newtonian flows. 
Results with a longer input window are reported in the Supplementary Information (Fig.~S1) and are used only to motivate this main two-snapshot setting.

\begin{figure*}[t]
    \centering
    \includegraphics[width=\textwidth]{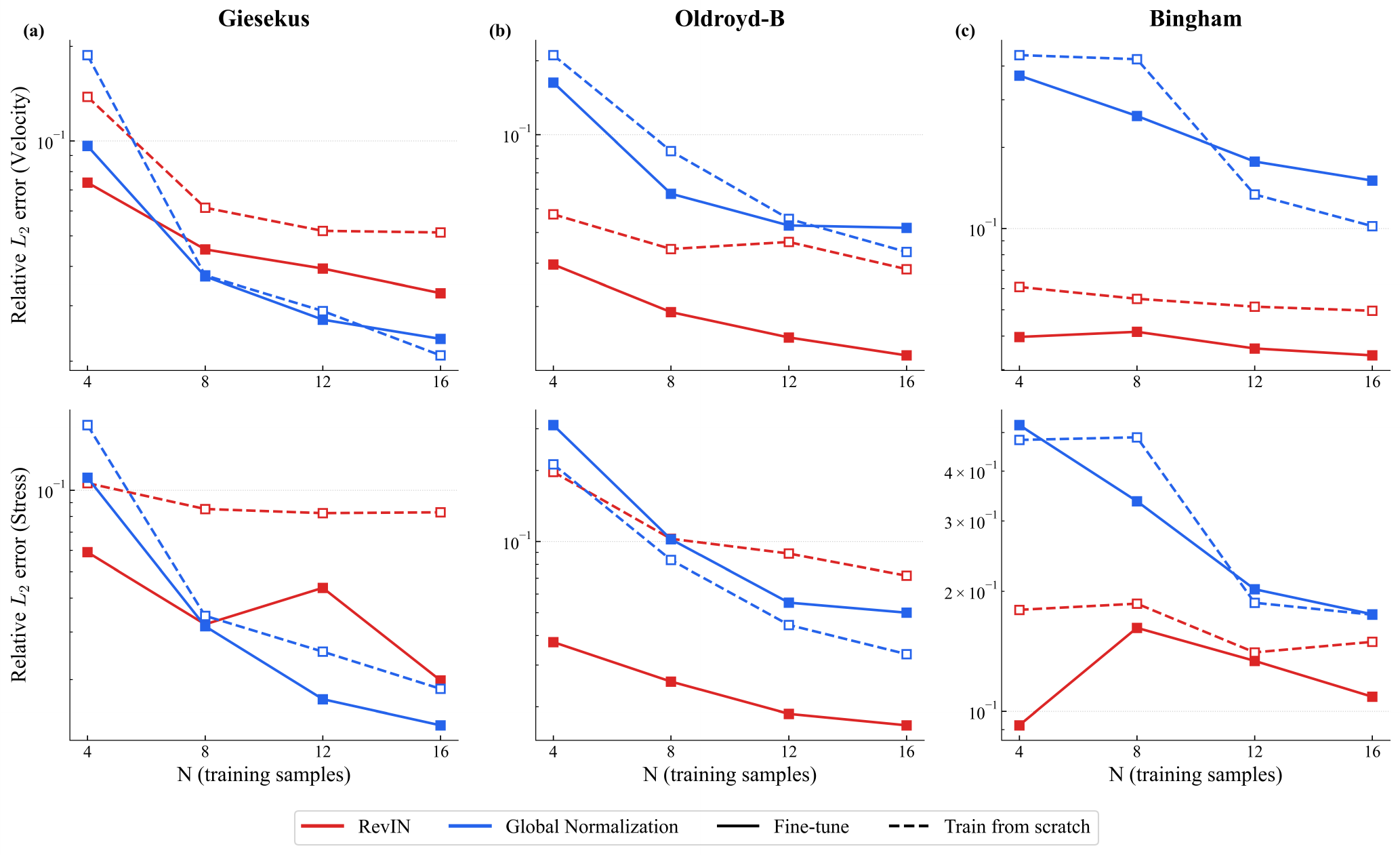}
    \caption{\textbf{Newtonian pretraining improves data-efficient prediction of non-Newtonian contraction flows.} Real-space relative $L_2$ velocity (top row) and stress (bottom row) error as a function of the number of non-Newtonian fine-tuning samples, $N$, for Giesekus, Oldroyd-B, and Bingham fluids in the $4{:}1$ contraction geometry. Global Normalization and RevIN are distinguished by color, while pretrained models and from-scratch models are shown by solid and dashed curves, respectively.}
    \label{fig:contraction_l2}
\end{figure*}

Before quantifying errors over the full validation set, Fig.~\ref{fig:contraction_fields} illustrates the predicted flow for representative cases with only $N=4$ target samples under RevIN, at nearly the same Reynolds number for the three fluids ($Re\approx16$). For each fluid, the ground-truth velocity magnitude at $t=5~\mathrm{s}$ is compared with the fine-tuned and from-scratch predictions and with the Newtonian pretraining simulation at the same $Re$. The largest differences between the models occur near the contraction, where the flow accelerates into the narrow channel and the constitutive response is most strongly expressed; far downstream, all predictions relax toward nearly the same developed profile. The profiles at $x=1$ show that the fine-tuned model follows the ground truth closely for all three fluids, whereas the from-scratch model trained on the same four samples visibly deviates. For Oldroyd-B and Bingham flow, the from-scratch prediction underestimates the centerline velocity ($2.10$ and $2.14~\mathrm{cm/s}$, compared with $2.32~\mathrm{cm/s}$ in both ground truths), while the fine-tuned predictions reach $2.26$ and $2.30~\mathrm{cm/s}$. For Giesekus flow, the from-scratch model misrepresents the near-wall profile, and the fine-tuned prediction retains the shape of the target profile. Over the plotted region, the relative $L_2$ error of $|\mathbf{v}|$ is reduced by pretraining from $13\%$ to $6\%$ for Giesekus, from $7\%$ to $3\%$ for Oldroyd-B, and from $8\%$ to $4\%$ for Bingham flow. The normalized error maps in Fig.~S2 of the Supplementary Information show that the from-scratch errors are concentrated at the re-entrant corner and along the downstream wall, where they reach up to $40\%$ of the maximum velocity, about $2$ times the largest errors of the fine-tuned predictions. 
The difference is even more pronounced in the shear stress, which is shown for a second set of representative cases at $Re\approx10$ in Fig.~S3. 
For Giesekus flow, the viscoelastic effect reduces the downstream shear stress by about an order of magnitude relative to the Newtonian flow at the same $Re$; the fine-tuned model reproduces this reduced stress level, with a relative $L_2$ error of $18\%$, compared with $48\%$ for the from-scratch model. 
The corresponding errors are $8\%$ and $20\%$ for Oldroyd-B and $6\%$ and $19\%$ for Bingham flow. 
Thus, with the same few target samples, the pretrained model recovers flow and stress features that the from-scratch model misses, and it adapts to the target fluid rather than reproducing the Newtonian solution on which it was pretrained.
These errors are obtained with the smallest fine-tuning budget of only 4 non-Newtonian samples and decrease as more target data are added (Fig.~\ref{fig:contraction_l2}); they nonetheless show that, even with such a small dataset, pretraining enables accurate predictions.

With this short-observation test in place, we first ask whether Newtonian pretraining produces a measurable advantage over training directly on the few available non-Newtonian samples.
Throughout the contraction-flow results, we compare two normalization choices: Global normalization and RevIN. The primary metric is the real-space relative $L_2$ error, reported for velocity and stress fields separately. 

Figure~\ref{fig:contraction_l2} reports the real-space relative $L_2$ velocity and stress errors as a function of the number of fine-tuning samples. 
The vertical separation between the solid and dashed curves represents the contribution of Newtonian pretraining at a fixed normalization method and data budget.
The benefit is most consequential in the low-data regime, where the pretrained model is adapted using only a few non-Newtonian simulations.
The clearest and most consistent separation appears under RevIN normalization: across all three constitutive models, the pretrained model remains below the corresponding from-scratch baseline throughout the few-shot regime.
Using the improvement metric defined in Sec.~\ref{sec:evaluation}, the RevIN results show positive transfer for every fluid, field grouping, and value of $N$ tested, with stress gains of 44--76\% for Giesekus, 75--81\% for Oldroyd-B, and 5--49\% for Bingham across the four fine-tuning budgets.

This trend supports the transfer hypothesis introduced above. In the two-snapshot setting, the model has only limited access to the case-specific temporal history, so the reduction in error cannot be attributed simply to extrapolating a long observed trajectory. Instead, the pretrained model appears to use information learned from the Newtonian contraction flows to organize the downstream non-Newtonian prediction. The transferred information is not the Newtonian constitutive law itself, but the geometry and momentum-transport structure that remains relevant after the stress response is changed.

The practical consequence is that Newtonian pretraining substantially reduces the amount of target data required to reach a given accuracy. 
In several cases, a pretrained model fine-tuned with only $4$ non-Newtonian simulations reaches an error comparable to or lower than a from-scratch model trained with $16$ simulations. 
For example, in Giesekus flow, the pretrained RevIN model at $N=4$ reaches a stress error of $0.059$, below the from-scratch error of $0.083$ at $N=16$.
Similarly, for Bingham flow, the pretrained RevIN model at $N=4$ reaches a stress error of $0.092$, compared with $0.149$ for the from-scratch model at $N=16$.
Therefore, in the short-observation regime, Newtonian pretraining provides information that is not recovered simply by moderately increasing the number of few-shot target examples.

These results establish the main contraction-flow transfer result: in the regime where the model cannot rely on a long input history, Newtonian pretraining provides a reusable representation for few-shot non-Newtonian prediction. 
Because the two training modes differ only in their initialization, this benefit is attributable to information acquired during Newtonian pretraining rather than to changes in architecture, optimization, or target-data exposure.
The reliability of this benefit, however, depends on how the fields are scaled. 
The RevIN curves show consistent transfer across all three fluids, while the Global Normalization curves in Fig.~\ref{fig:contraction_l2} are more variable. 
This difference is analyzed next.

\subsection{Normalization controls transfer reliability}
\label{sec:results_normalization}

The transfer benefit established above is not a property of pretraining alone; it also depends on how the physical fields are scaled before they are presented to the model. 
Under RevIN, pretraining gives reliable improvement across all three fluids, whereas under Global normalization the effect is weaker, more variable, and in some cases negative. 
This does not contradict the transfer result above; instead, it shows that the transferred representation must be paired with an input formulation that preserves the relevant velocity and stress scales during fine-tuning.

The contrast is clearest in the stress errors. 
With Global Normalization, Giesekus still benefits from pretraining, with stress error reductions ranging from $8$ to $36\%$ across the four values of $N$. 
Oldroyd-B shows the opposite behavior: the pretrained model is not better than the from-scratch model at any fine-tuning budget under Global normalization. 
Bingham is intermediate and non-monotonic, with Global normalized stress improvements changing sign across $N$.
Thus, Global Normalization can preserve useful pretraining gains in some datasets, but it can also suppress or reverse them in others.
Pretraining therefore does not provide a normalization-independent advantage; its effectiveness depends on whether the input formulation preserves the relevant field scales during fine-tuning.

The same dependence appears in the absolute errors, and it is not uniform across fluids. 
For Oldroyd-B and Bingham, RevIN gives the lowest error at every value of $N$, often by a wide margin. 
At $N=16$, the Oldroyd-B stress error is $0.017$ with RevIN compared with $0.050$ with Global normalization. 
For Bingham, the corresponding stress errors are $0.109$ and $0.175$, respectively. 
The difference is even larger for Bingham velocity in the most data-limited case, where the pretrained RevIN error at $N=4$ is $0.040$ compared with $0.369$ under Global normalization. 
For Giesekus, however, the ordering reverses beyond the smallest budget: RevIN is more accurate at $N=4$, but Global normalization achieves lower velocity and stress errors for $N\ge 8$. 
At $N=16$, the Giesekus stress error is $0.014$ with Global normalization and $0.020$ with RevIN.

This dataset dependent behavior is consistent with the dynamic range of the field magnitudes in each dataset, which is the quantity that Global Normalization handles poorly and that RevIN is designed to remove. 
Global normalization applies a single mean and standard deviation, computed across the training set, to every sample. 
When a dataset spans a wide range of velocity and stress magnitudes, these shared statistics can be dominated by high-magnitude samples, causing low-magnitude samples or regions to be compressed in the normalized representation. 
RevIN avoids this by rescaling each sample with its own input statistics before the operator processes it and then mapping the prediction back to physical space. As a result, RevIN is better suited to datasets where velocity and stress magnitudes vary strongly from sample to sample or where physically meaningful near-zero regions must be resolved.
A second, related factor is the scale mismatch between the source and target data. Under Global normalization, the pretrained model retains the statistics of the Newtonian pretraining data, so the non-Newtonian fields are presented to it on a scale that differs from the one seen during pretraining. This mismatch is largest for Oldroyd-B, whose normal-stress standard deviation is about $12$ times that of the Newtonian data. RevIN removes this mismatch because the statistics are recomputed for every sample.

\begin{table}[t]
\centering
\caption{Datasets for the contraction-flow experiments.}
\label{tab:contraction_datasets}
\begin{tabular}{lcc}
\hline
Fluid & Reynolds number & Additional parameter \\
\hline
Newtonian & $Re\in[0.4,80]$ & $Bi=0, Wi=0$ \\
Giesekus & $Re\in[1,42]$ & $Wi=0.1\,Re$ \\
Oldroyd-B & $Re\in[0.4,80]$ & $Wi=0.01\,Re$ \\
Bingham & $Re\in[0.4,60]$ & $Bi\in[0,5]$ \\
\hline
\end{tabular}
\end{table}

The three datasets differ markedly in this respect (Table~\ref{tab:contraction_datasets}). Oldroyd-B spans the widest Reynolds-number interval, with $Re\in[0.4,80]$, and Bingham combines a wide Reynolds-number range with an independent yield-stress variation and near-zero velocities in the unyielded regions ($Re\in[0.4,60]$, $Bi\in[0,5]$). 
Both therefore contain the wide magnitude variation where a single set of global statistics is inadequate and underrepresents the low-amplitude but physically important structures.

Giesekus provides the complementary case where data samples occupy a comparatively narrower range of field magnitudes and do not contain the small magnitude velocity ranges ($Re < 1$) that challenge global scaling in Bingham flow. Consequently, global statistics remain representative of each sample and Global Normalization stays competitive. 
This behavior should therefore not be read as an exception to the normalization result, but as confirmation of the range-based interpretation: RevIN is most valuable when sample-wise scaling resolves variation that would otherwise be compressed by dataset-level statistics.

Taken together, these results refine the main finding. Newtonian pretraining supplies reusable structure for few-shot non-Newtonian prediction, but realizing that benefit reliably across constitutive models requires a normalization that is robust to the per-sample magnitude variation of the target dataset. 
RevIN provides such a normalization and makes the pretraining benefit reliable for the wide-range cases; Global Normalization remains sufficient only when the dataset is narrow enough in dynamic range that its fixed statistics remain representative. Having identified this scaling dependence, we next move to the oscillatory wake case.

\begin{figure}[t]
    \centering
    \includegraphics[width=8.7cm]{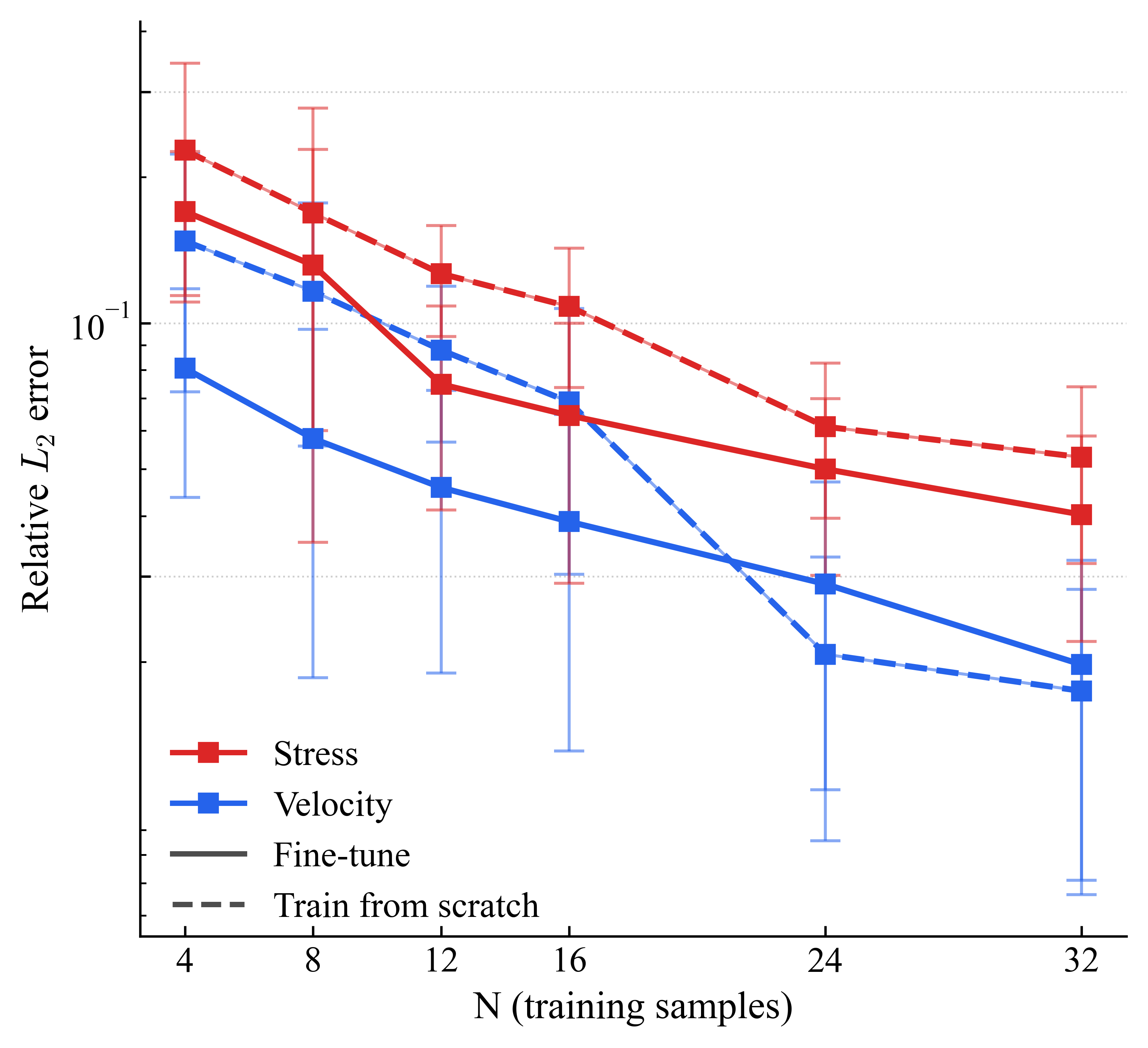}
    \caption{\textbf{Prediction error in the triangular-obstacle flow.} Real-space relative $L_2$ error as a function of the number of fine-tuning samples, $N$, under Global normalization. Solid curves denote Newtonian-pretrained models after fine-tuning, and dashed curves denote models trained from scratch using the same target samples and training protocol. Red and blue curves correspond to stress and velocity errors, respectively, and error bars denote the standard deviation across validation samples.}
    \label{fig:triangle_l2}
\end{figure}

\subsection{Wake dynamics transfer in an oscillatory flow}
\label{sec:results_triangle}

The contraction-flow results establish that Newtonian pretraining improves few-shot non-Newtonian prediction when the pretraining and downstream data are aligned in geometry and the flow response is dominated by contraction-induced structures. 
We next test a qualitatively different regime: Oldroyd-B flow past a triangular obstacle. 
In this geometry, the downstream wake develops periodic vortex shedding, so successful prediction requires more than reconstructing a spatial field at isolated times; the model must also reproduce the temporal dynamics of the wake.

The triangle experiment therefore serves a different role from the contraction study. 
The contraction cases test transfer in a short-observation setting, whereas the triangle wake tests whether the transferred representation can recover sustained unsteady dynamics. 
For this reason, the triangle experiments use Global Normalization in the main text and a longer input window of 10 snapshots ($t\in[0,1.8]~\mathrm{s}$), as defined in Sec.~\ref{sec:problem_transfer}. 
This longer window provides an initial history of the wake state before the model predicts the remaining trajectory to $t=10~\mathrm{s}$. 
The central question is therefore not only whether pretraining lowers the pointwise relative $L_2$ error, but whether the resulting prediction contains the correct physics and the vortex-shedding response.

We first evaluate the triangle wake using the same real-space relative $L_2$ metric used in the contraction-flow experiments. Figure~\ref{fig:triangle_l2} shows that Newtonian pretraining gives a clear pointwise advantage in the few-shot regime under Global normalization. 
Qualitatively, the pretrained stress curve remains below the from-scratch curve across all training budgets, indicating a persistent transfer benefit for stress prediction. 
For velocity, the pretrained model is more accurate at small and intermediate budgets, while the from-scratch model becomes lower at the two largest budgets. 
This pattern supports the same data-efficiency conclusion observed in contraction flows, especially when only a few target simulations are available, but it also shows that the high-budget velocity comparison is more ambiguous.

Pretraining reduces the stress error at every value of $N$, with improvements ranging from $18$ to $41\%$ across the six fine-tuning budgets. 
For velocity, the pretrained model reduces the error by $43$--$50\%$ for $N\le16$, after which the from-scratch model becomes lower at $N=24$ and $N=32$. 
The pointwise metric therefore establishes a few-shot transfer benefit and a persistent stress advantage, but it does not by itself determine whether the oscillatory wake dynamics have been recovered correctly. 

\begin{figure}[t]
    \centering
    \includegraphics[width=8.7cm]{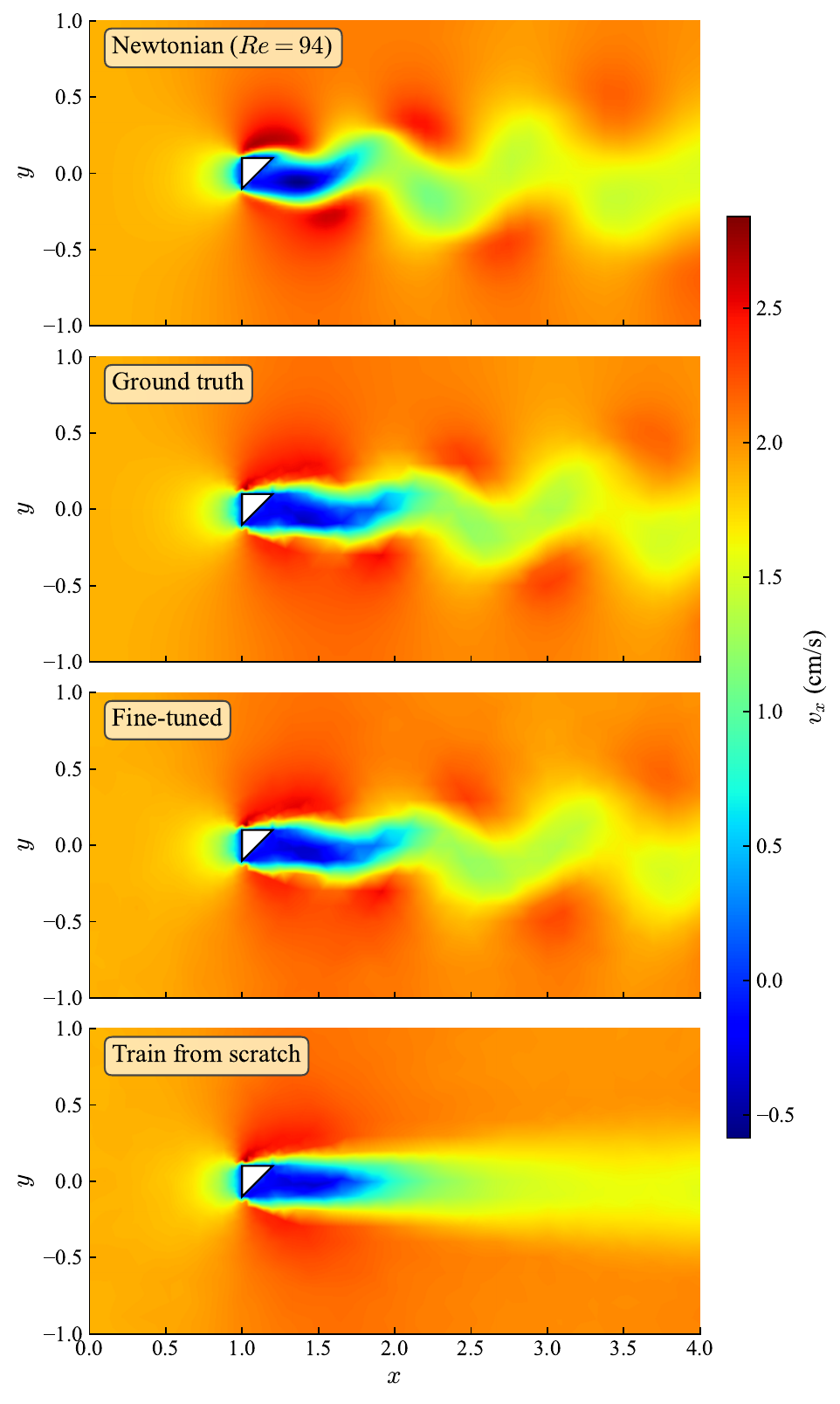}
    \caption{\textbf{Representative wake fields.} Newtonian simulation at the same Reynolds number, ground truth, pretrained (fine-tuned) prediction, and from-scratch prediction of the streamwise velocity field, $v_x$, for a representative triangular-obstacle Oldroyd-B case at $Re = 94$ and $Wi = 0.94$ using $N=12$ training samples under Global Normalization.}
    \label{fig:triangle_fields}
\end{figure}

To understand what this error comparison corresponds to physically, we next inspect representative wake fields.
Figure~\ref{fig:triangle_fields} shows the streamwise velocity field $v_x$ for a representative case using only $N=12$ training samples. 
The ground truth contains a wavy downstream wake characteristic of vortex shedding. 
The pretrained model reproduces this alternating wake structure, including the downstream spatial modulation. 
In contrast, the from-scratch model predicts a smooth and nearly steady wake, with no oscillations. 
This visual comparison suggests that the difference between the models is not only a difference in pointwise error, but also a difference in the recovered unsteady wake response.
The top panel of Fig.~\ref{fig:triangle_fields} shows the Newtonian simulation at the same Reynolds number. Its recirculation region behind the obstacle is shorter, and its wake begins to undulate closer to the obstacle than in the Oldroyd-B ground truth. The pretrained prediction follows the Oldroyd-B wake rather than the Newtonian one, indicating that fine-tuning adapts the transferred representation to the viscoelastic target instead of reproducing the Newtonian source flow.

\begin{figure}[t]
    \centering
    \includegraphics[width=8.7cm]{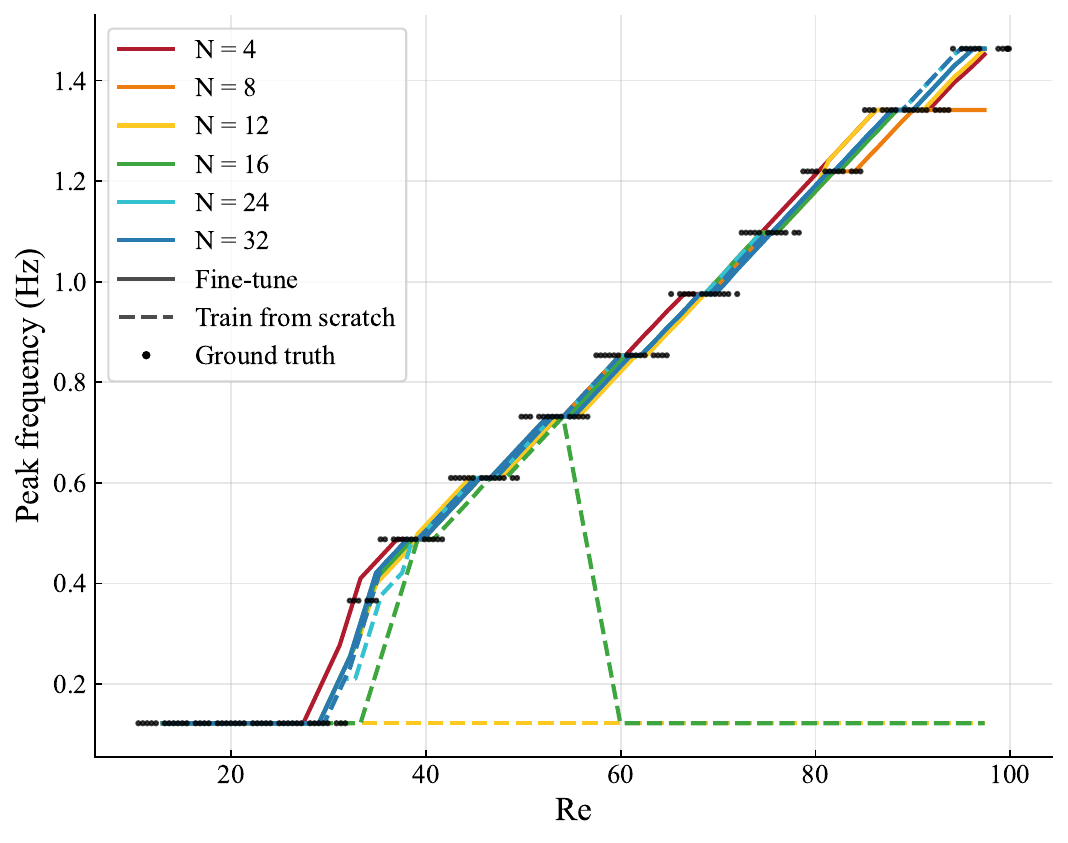}
    \caption{\textbf{Pretraining recovers the shedding-frequency trend.} Dominant temporal frequency in the wake as a function of Reynolds number for the triangular-obstacle Oldroyd-B flow under Global normalization. Ground-truth frequencies are shown as reference markers, and model predictions are compared for pretrained and from-scratch models across fine-tuning budgets. The pretrained models track the increasing frequency--$Re$ trend, whereas the from-scratch models trained with few samples show a weaker or nearly flat frequency response.}
    \label{fig:triangle_frequency}
\end{figure}

\begin{figure}
    \centering
    
    \includegraphics[width=8.7cm]{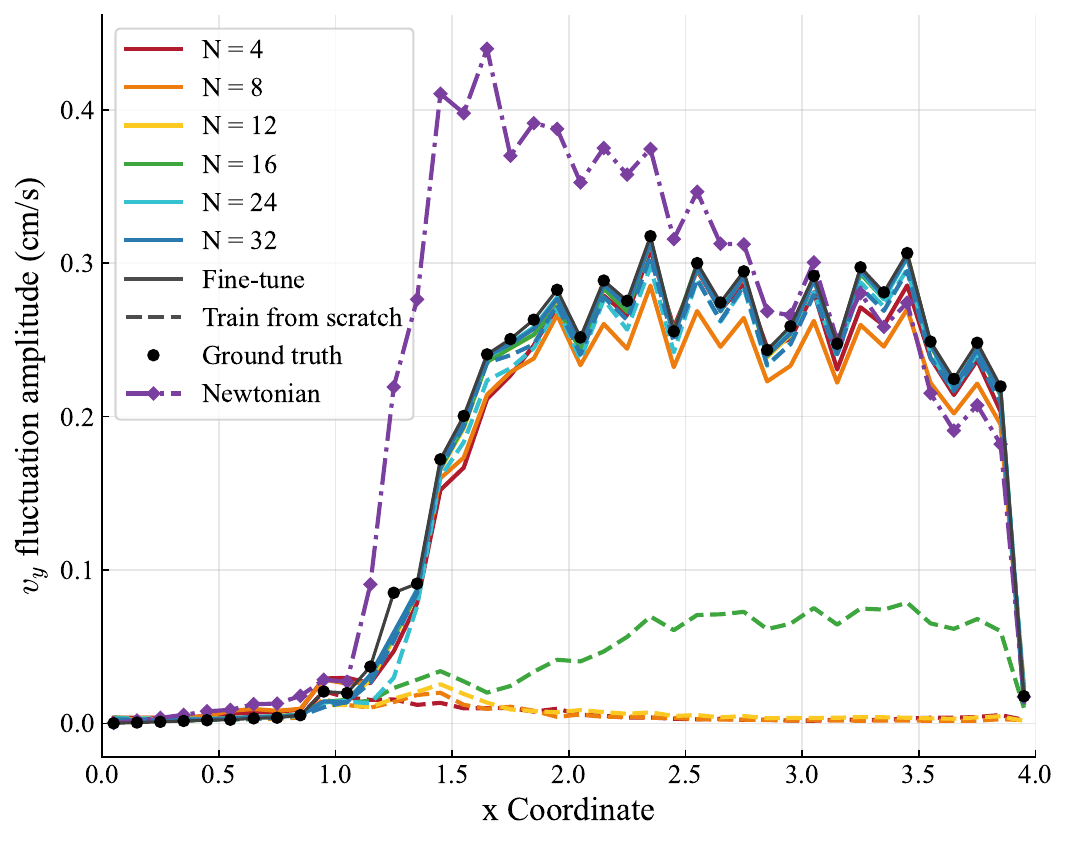}
    \caption{\textbf{Pretraining recovers the wake-fluctuation amplitude.} RMS fluctuation amplitude of the transverse velocity, $v_y$, measured in the downstream wake for a representative triangular-obstacle case at $Re = 94$ and $Wi = 0.94$ under Global normalization, as a function of the streamwise coordinate measured from the channel inlet.
    Colors denote the number of non-Newtonian fine-tuning samples, $N$, solid curves denote Newtonian-pretrained models after fine-tuning, dashed curves denote models trained from scratch, and black markers denote the ground truth.
    The purple dash-dotted curve shows the Newtonian simulation at the same Reynolds number.
    The RMS envelope compares the strength and spatial persistence of the oscillatory wake in the ground truth, pretrained, and from-scratch prediction across different training budgets.
    \label{fig:triangle_rms}}
\end{figure}

Although the field comparison provides a useful visual interpretation, a single snapshot and snapshot-wise error metric cannot verify the temporal content of an oscillatory flow.
A model can obtain a relatively low pointwise error by predicting a smooth wake close to the time-averaged solution, even if it weakens or suppresses vortex shedding. 
Conversely, a prediction that contains the correct unsteady structure can still be penalized if the vortices are slightly shifted in phase, amplitude, or spatial position relative to the reference. 
Relative $L_2$ error is therefore useful for measuring instantaneous field agreement, but it must be complemented by phase-insensitive diagnostics that directly test the temporal content of the wake.

We next examine the dominant shedding frequency, which directly measures whether the model recovers the temporal scale of the wake. Figure~\ref{fig:triangle_frequency} compares the peak temporal frequency of $v_x$ extracted from the wake as a function of Reynolds number.
Below $Re\approx30$, the ground-truth peak remains at the lowest resolved frequency, indicating that no shedding is resolved over the prediction window. Above this onset, the ground-truth data show an increasing peak frequency with Reynolds number, as expected for stronger and faster vortex shedding at higher flow rates.
The pretrained models follow this frequency--$Re$ trend, including the shedding onset, across the tested fine-tuning budgets, whereas the from-scratch models trained with $N\le12$ remain at the lowest resolved frequency over the entire $Re$ range, i.e., they do not produce a resolved oscillation. The from-scratch model at $N=16$ follows the trend only up to $Re\approx50$, and only the from-scratch models at $N=24$ and $N=32$ recover the ground-truth trend.
Quantitatively, among the $127$ validation cases with resolved shedding, the pretrained models recover the ground-truth peak-frequency bin in $87$--$95\%$ of the cases at every budget, whereas the from-scratch models recover it in none of the cases for $N\le12$ and in $26\%$ of the cases at $N=16$. 
This comparison supports the interpretation suggested by the field snapshots: the pretrained model is not only reducing pointwise error in the few-shot regime, but also recovering the characteristic time scale of vortex shedding.

Frequency verifies the time scale of the oscillation, but not its strength. 
We therefore also examine the wake-fluctuation amplitude using the RMS fluctuation of the transverse velocity, $v_y$, along the downstream wake. 
This metric removes the mean flow and measures how strongly the unsteady component persists behind the obstacle. 
As shown in Fig.~\ref{fig:triangle_rms}, the ground-truth fluctuation amplitude rises sharply downstream of the obstacle and remains large throughout the wake. 
The Newtonian-pretrained models reproduce this envelope across all fine-tuning budgets, with only a slight underestimation for $N=4$ and $N=8$, closely tracking both the sharp rise in fluctuations near $x\approx1.4$ and the downstream fluctuation level.
In contrast, the from-scratch models trained with fewer target samples strongly underestimate the unsteady fluctuation amplitudes. 
For $N=4$, $8$, and $12$, the RMS amplitude collapses toward a nearly steady wake and remains close to zero over most of the wake region.
The from-scratch model at $N=16$ recovers some fluctuation growth, but its downstream amplitude remains substantially smaller than the ground truth, reaching only about $0.05$--$0.08$~cm/s in regions where the reference amplitude is roughly $0.25$--$0.30$~cm/s.
At larger budgets, the from-scratch predictions improve and approach the ground-truth level, consistent with the high-budget velocity behavior in Fig.~\ref{fig:triangle_l2}. 
The Newtonian reference at the same Reynolds number differs clearly from the Oldroyd-B ground truth: its fluctuations rise earlier, near $x\approx1.2$, and reach a peak amplitude of about $0.44~\mathrm{cm/s}$, compared with about $0.32~\mathrm{cm/s}$ for the viscoelastic wake. The fine-tuned models follow the weaker Oldroyd-B envelope rather than the Newtonian one, showing that the recovered fluctuation strength reflects the viscoelastic target and not the Newtonian flow seen during pretraining.
The RMS comparison therefore shows that pretraining recovers the wake-fluctuation strength much more reliably in the few-shot regime.
Together with the shedding-frequency results, this indicates that pretraining recovers both the temporal scale and the fluctuation amplitude of vortex shedding, rather than merely lowering a pointwise error.

\section{Conclusion}
\label{sec:conclusion}

We introduced RheOPT, a framework that pretrains a transformer-based neural operator on inexpensive Newtonian simulations and then fine-tunes it, using only a small number of non-Newtonian samples, into a surrogate for complex-fluid flow. 
The central question was whether a representation learned under a Newtonian stress response remains useful after the constitutive law changes. 
We examined this question in two geometry-matched settings: a $4{:}1$ contraction channel for Giesekus, Oldroyd-B, and Bingham fluids, and the oscillatory wake behind a triangular obstacle for Oldroyd-B flow. 
Because the pretraining and fine-tuning data share the same domain in each setting, the experiments isolate cross-constitutive transfer: the model must adapt from Newtonian momentum transport to viscoelastic or viscoplastic stress evolution without also extrapolating to a new geometry.

The contraction experiments provide the main evidence for this transfer. 
In a deliberately strict two-snapshot input setting, Newtonian pretraining consistently improves few-shot prediction of non-Newtonian flow when the transferred representation is paired with an appropriate normalization. 
Under RevIN, the pretrained model outperforms the corresponding from-scratch baseline across the tested fluids, output groupings, and target-data budgets, and the benefit is most consequential in the low-data regime. 
In several cases, a pretrained model fine-tuned on 4 non-Newtonian simulations reaches an error comparable to or lower than a from-scratch model trained on 16 simulations. 
Because the pretrained and from-scratch models use the same architecture, optimizer, training schedule, target samples, and seed protocol, and differ only in their initialization, this gain is attributable to information acquired during Newtonian pretraining. 
The transferred information is not the Newtonian constitutive response itself, but the geometry-conditioned momentum-transport structure that remains useful after the stress model changes.
Consistently, representative fields show that the fine-tuned model reproduces the target flow rather than the Newtonian source flow at the same Reynolds number.

The results also show that this transfer is not independent of the input representation. 
RevIN, which rescales each sample using its own statistics and then restores the physical scale after prediction, delivers reliable transfer across all three fluids.
Global normalization, by contrast, is dataset dependent and can weaken or reverse the gain. We traced this behavior to the dynamic range of each dataset and to the scale mismatch between the Newtonian source statistics and the non-Newtonian target fields.
When velocity and stress magnitudes vary widely across samples, fixed dataset-level statistics can compress low-magnitude but physically important structures. 
Per-sample normalization reduces this scale imbalance and allows the pretrained representation to be used more effectively during fine-tuning. 
Thus, Newtonian pretraining supplies reusable structure, but the normalization determines how reliably that structure is expressed in the downstream non-Newtonian model.

The triangular-obstacle wake extends the analysis from contraction-dominated flow to sustained unsteady dynamics. 
In this oscillatory setting, pointwise relative $L_2$ error provides a useful first comparison, but it is not sufficient to verify that the wake dynamics have been recovered. 
A smooth prediction can remain close to the time-averaged wake while suppressing vortex shedding, while a prediction with a dynamically correct shedding structure but a small phase offset can be penalized.
We therefore complemented pointwise error with phase-insensitive metrics. 
The pretrained model follows the dominant shedding-frequency trend with Reynolds number and preserves wake fluctuation amplitudes more closely than the from-scratch model. 
Representative velocity fields show the same behavior: the pretrained model reproduces the oscillatory downstream wake, whereas the from-scratch model trained on the same samples largely misses this unsteady response. 
These results show that the transferred representation can support not only pointwise accuracy in the few-shot regime, but also recovery of physically meaningful wake dynamics.

Several limitations follow from the design of the study. 
The experiments here are geometry matched by construction. 
This choice isolates transfer across constitutive laws, but it leaves open whether the same pretrained representation can transfer when the geometry also changes between pretraining and fine-tuning. 
Additionally, the normalization results show a clear empirical pattern, but a more systematic theory of how field scaling interacts with operator pretraining remains open. 
Hence, future extensions can include transfer across geometries, additional constitutive models, broader parameter ranges, three-dimensional flows, and adaptation strategies that better preserve the pretrained dynamics during fine-tuning. 

Overall, RheOPT shows that Newtonian simulations can serve as a physically grounded pretraining source for data-scarce non-Newtonian prediction. 
The results support a simple interpretation: Newtonian pretraining does not teach the model the downstream constitutive law, but it does provide a useful representation of geometry-driven transport that can be adapted with few target simulations. 
Together, these findings establish Newtonian pretraining as a practical route toward more data-efficient neural-operator surrogates for complex-fluid flows.

\begin{acknowledgments}
M.S. and S.J. would like to acknowledge the support of DoD MURI Award N00014-23-1-2499.

\end{acknowledgments}

\section*{Data Availability Statement} 
The code for RheOPT, including hyperparameter configurations, the simulated datasets, training scripts and evaluation pipelines, is publicly available in the GitHub repository at \url{https://github.com/procf}.



\bibliography{apssamp}
\end{document}


\title{Supplementary Information\\
RheOPT: Few-Shot Non-Newtonian Flow Prediction \\
via Newtonian Pretraining of RheOFormer}

\author{Maedeh Saberi}
\affiliation{\mbox{Department of Mechanical and Industrial Engineering, Northeastern University, Boston, Massachusetts 02115, USA}}
\author{Amir Barati Farimani}
\affiliation{Department of Mechanical Engineering, Carnegie Mellon University, Pittsburgh, Pennsylvania 15213, USA}
\author{Safa Jamali}
 \email{s.jamali@northeastern.edu}
\affiliation{\mbox{Department of Mechanical and Industrial Engineering, Northeastern University, Boston, Massachusetts 02115, USA}}

\maketitle

This Supplementary Information provides the constitutive relations of the target fluids (Sec.~\ref{si:constitutive}), the parameter ranges, sampled values, and dataset partitions (Sec.~\ref{si:datasets}), the computational domains, meshes, and numerical settings (Sec.~\ref{si:meshes}), the architectural and training hyperparameters of RheOFormer and RheOPT (Sec.~\ref{si:hyper}), and the contraction-flow results obtained with a longer input window (Sec.~\ref{si:longwindow}).

\section{Constitutive relations}
\label{si:constitutive}

All fluids satisfy the incompressibility constraint and momentum balance of the main text,
\begin{equation}
\nabla\cdot\mathbf{u}=0,\qquad
\rho\left(\frac{\partial\mathbf{u}}{\partial t}+\mathbf{u}\cdot\nabla\mathbf{u}\right)=\nabla\cdot\boldsymbol{\sigma},\qquad
\boldsymbol{\sigma}=-p\mathbf{I}+\boldsymbol{\tau},
\end{equation}
and differ only in the constitutive relation for the non-hydrostatic stress $\boldsymbol{\tau}$.
The stress channels used as model inputs and outputs are the components of the total stress $\boldsymbol{\sigma}$, and $\mathbf{D}=(\nabla\mathbf{u}+\nabla\mathbf{u}^{T})/2$ denotes the rate-of-deformation tensor.

\paragraph*{Newtonian fluid.}
For the Newtonian source simulations, $\boldsymbol{\tau}=2\eta\mathbf{D}$ with a constant viscosity $\eta$.

\paragraph*{Oldroyd-B fluid.}
The stress is decomposed into a Newtonian solvent contribution and a polymeric contribution, $\boldsymbol{\tau}=2\eta_s\mathbf{D}+\boldsymbol{\tau}_p$, where the polymeric stress satisfies
\begin{equation}
\boldsymbol{\tau}_p+\lambda\,\overset{\triangledown}{\boldsymbol{\tau}}_p=2\eta_p\mathbf{D},
\end{equation}
with
\begin{equation}
\overset{\triangledown}{\boldsymbol{\tau}}_p=\frac{\partial\boldsymbol{\tau}_p}{\partial t}+\mathbf{u}\cdot\nabla\boldsymbol{\tau}_p-(\nabla\mathbf{u})^{T}\cdot\boldsymbol{\tau}_p-\boldsymbol{\tau}_p\cdot\nabla\mathbf{u}.
\end{equation}
Here, $\lambda$ is the relaxation time, $\eta_p$ is the polymer viscosity, and $\overset{\triangledown}{\boldsymbol{\tau}}_p$ is the upper-convected derivative.
The total zero-shear viscosity is $\eta_0=\eta_s+\eta_p$, and the viscosity ratio is $\beta=\eta_s/\eta_0$.
In the contraction dataset, the ratio of the solvent to the total shear stress in the fully developed downstream channel is $0.10$ for all cases, consistent with $\beta=1/9$.

\paragraph*{Giesekus fluid.}
The Giesekus model adds a quadratic stress term that introduces shear thinning and a nonzero second normal-stress difference,
\begin{equation}
\boldsymbol{\tau}_p+\lambda\,\overset{\triangledown}{\boldsymbol{\tau}}_p+\frac{\alpha\lambda}{\eta_p}\,\boldsymbol{\tau}_p\cdot\boldsymbol{\tau}_p=2\eta_p\mathbf{D},
\end{equation}
with the mobility parameter $\alpha$ and viscosity ratio.

\paragraph*{Bingham fluid.}
The Bingham fluid behaves as a rigid material below the yield stress $\tau_y$ and flows with plastic viscosity $\mu_p$ above it,
\begin{equation}
\boldsymbol{\tau}=\left(\mu_p+\frac{\tau_y}{\dot{\gamma}}\right)2\mathbf{D}\quad\text{if}\quad|\boldsymbol{\tau}|>\tau_y,
\qquad
\mathbf{D}=\mathbf{0}\quad\text{otherwise},
\end{equation}
where $\dot{\gamma}=\sqrt{2\,\mathbf{D}:\mathbf{D}}$ and $|\boldsymbol{\tau}|=\sqrt{\boldsymbol{\tau}:\boldsymbol{\tau}/2}$.
The apparent viscosity in the simulations ranges from $\mu_p$ to approximately $5.6\times10^{4}\,\mu_p$.

\paragraph*{Dimensionless numbers.}
The Reynolds, Weissenberg, and Bingham numbers are defined as
\begin{equation}
Re=\frac{\rho U L}{\eta_0},\qquad
Wi=\frac{\lambda U}{L},\qquad
Bi=\frac{\tau_y L}{\mu_p U},
\end{equation}
where, for the contraction datasets, $U=4U_{in}$ is the mean velocity in the downstream channel, $U_{in}$ is the mean inlet velocity, and $L=0.01~\mathrm{m}$ is the half-width of the downstream channel. The simulation file names and metadata use the inlet velocity instead; the values reported here and in the main text are therefore $Re=4\,Re_{\mathrm{file}}$, $Wi=4\,Wi_{\mathrm{file}}$, and $Bi=Bi_{\mathrm{file}}/4$. All contraction datasets, including the Newtonian source, use the same fluid density, viscosity, and geometry.
The fluid density is $\rho=1000~\mathrm{kg\,m^{-3}}$, and the zero-shear (or plastic) viscosity is $0.01~\mathrm{Pa\,s}$; for the Bingham fluid, $\eta_0$ in $Re$ is replaced by $\mu_p$.
In the Giesekus and Oldroyd-B contraction datasets, the inlet velocity is the only varied control, so $Re$ and $Wi$ change together, with $Wi/Re=0.1$ for Giesekus and $Wi/Re=0.01$ for Oldroyd-B.
With these values, the relations correspond to $\lambda=1~\mathrm{s}$ for Giesekus and $\lambda=0.1~\mathrm{s}$ for Oldroyd-B.

\section{Parameter ranges, sampled values, and dataset partitions}
\label{si:datasets}

Table~\ref{tab:si_datasets} summarizes the source and target datasets.
For each target fluid, the simulations are split into a fine-tuning pool and a fixed validation set.
The fine-tuning samples are drawn from the pool for each budget $N$, and the same validation set is used for all budgets, seeds, and training modes of that fluid.
The sampled cases used for fine-tuning are listed in Table~\ref{tab:si_trainsets}.
The Newtonian source datasets are used in full for pretraining.

All simulations are saved at $\Delta t=0.1~\mathrm{s}$ and subsampled to $\Delta t=0.2~\mathrm{s}$ for training and evaluation.
For the contraction datasets, the first two subsampled snapshots form the model input, so that the predicted horizon is $t\in[0.4,5]~\mathrm{s}$ for Oldroyd-B and Bingham and $t\in[0.4,8]~\mathrm{s}$ for Giesekus.
For the triangular obstacle, the first ten subsampled snapshots form the input, and the predicted horizon is $t\in[2,10]~\mathrm{s}$.
The Bingham dataset combines $11$ Reynolds numbers, $Re\in\{0.4,0.6,1,2,3,4,8,16,28,40,60\}$, with a subset of the yield-stress levels $Bi\in\{0,\,0.0125,\,0.025,$ $\,0.0625,\,0.125,\,0.25,\,0.5,\,1,\,1.5,\,2,\,2.5,\,3,\,3.75,\,5\}$.
The Newtonian contraction data are stored with velocities in $\mathrm{cm/s}$, whereas the non-Newtonian contraction data are stored in $\mathrm{m/s}$. In consistent units, all contraction datasets satisfy $U_{in}=2.5\times10^{-4}Re~\mathrm{m/s}$, and the inlet velocities measured directly from the stored fields agree with these values to within $0.5\%$.

\section{Computational domains, meshes, and numerical settings}
\label{si:meshes}

The $4{:}1$ planar contraction is simulated on the upper half of the channel, $x\in[-10,10]$ and $y\in[0,4]$, with a symmetry condition at $y=0$.
Coordinates are given in centimeters: the upstream channel has half-height $4~\mathrm{cm}$ for $x<0$, and the downstream channel has half-height $1~\mathrm{cm}$ for $x>0$, which is the length scale $L=0.01~\mathrm{m}$ used in $Re$, $Wi$, and $Bi$.
The triangular-obstacle domain spans $x\in[0,5]$ and $y\in[-1,1]$, with $x$ measured from the channel inlet.
The obstacle is a right triangle with vertices at $(1,\,0.1)$, $(1,\,-0.1)$, and $(1.2,\,0.1)$, and the figures in the main text show the region $x\in[0,4]$.
Table~\ref{tab:si_meshes} lists the mesh sizes.
The Newtonian source and non-Newtonian target simulations of each geometry share the same domain but use different unstructured meshes.
Because RheOFormer encodes and decodes the fields directly at the node coordinates, both meshes are used without interpolation.

\section{Architecture and training hyperparameters}
\label{si:hyper}

RheOPT uses the RheOFormer architecture of Ref.~\cite{saberi2025rheoformer_si} without modification.
Table~\ref{tab:si_hyper} lists the architectural and training hyperparameters.
The same architecture is used for pretraining, fine-tuning, and training from scratch, with approximately $6.8\times10^{6}$ trainable parameters.

\section{Contraction-flow results with a longer input window}
\label{si:longwindow}

To motivate the two-snapshot setting of the main text, we repeated the contraction experiments with a longer input window of ten snapshots, $t\in[0,1.8]~\mathrm{s}$.
The Newtonian pretraining and the non-Newtonian fine-tuning were both performed with this window, and all other settings were kept unchanged.
Figure~\ref{fig:si_longwindow} shows the resulting real-space relative $L_2$ errors, and Table~\ref{tab:si_longwindow} reports the corresponding values and improvements averaged over three seeds.

With a longer input history, both training modes reach considerably lower errors than in the two-snapshot setting, and the separation between the pretrained and from-scratch models becomes smaller.
Under RevIN, the pretraining benefit for Giesekus and Oldroyd-B is largest at $N=4$, with stress improvements of $16\%$ and $31\%$, and becomes negligible for $N\ge12$, where both models reach nearly the same error.
For Bingham with RevIN, pretraining increases the stress error at $N=4$ but reduces it by $20$--$31\%$ for $N\ge8$.
Under Global normalization, pretraining remains beneficial for all three fluids, most notably for the Bingham stress, with improvements of $68$--$77\%$ for $N\ge8$.
Thus, when the model observes a long initial trajectory, much of the information needed to predict the subsequent evolution is already contained in the input, and the from-scratch model can recover a large part of the accuracy without pretraining.
This motivates the strict two-snapshot setting used in the main text, in which the observed history is limited and the contribution of Newtonian pretraining can be assessed more directly.

\section{Representative contraction fields and error maps}
\label{si:fields}

Figures~\ref{fig:si_vmag} and \ref{fig:si_sxy} complement the representative velocity fields of the main text (Fig.~2) with error maps and with the corresponding shear-stress fields.
All predictions are obtained with $N=4$ fine-tuning samples under RevIN and the two-snapshot input, and are shown at $t=5~\mathrm{s}$.
The error maps show the signed difference between the prediction and the ground truth, normalized by the maximum of the ground-truth field of the same case, $(f-f_{\mathrm{GT}})/\max|f_{\mathrm{GT}}|$.
The two error rows of each column share one symmetric color scale, so that the fine-tuned and from-scratch errors can be compared directly.

\begin{table*}
\centering
\caption{Source and target datasets. For the contraction, $Re$, $Wi$, and $Bi$ follow the definitions of Sec.~\ref{si:constitutive}, with the downstream velocity $U=4U_{in}$ and $L=0.01~\mathrm{m}$.}
\label{tab:si_datasets}
\footnotesize
\begin{tabular}{llcccccc}
\hline
Dataset & Geometry & Cases & Validation & Parameter range & Inlet velocity & Snapshots & Saved time \\
\hline
Newtonian (source) & $4{:}1$ contraction & 200 & --- & $Re\in[0.4,80]$, $\Delta Re=0.4$ & $U_{in}=2.5\times10^{-4}Re~\mathrm{m/s}$ & 51 & $t\in[0,5]$ s \\
Giesekus & $4{:}1$ contraction & 28 & 12 & $Re\in[1,42]$, $Wi=0.1\,Re$ & $U_{in}=2.5\times10^{-4}Re~\mathrm{m/s}$ & 51 & $t\in[0,5]$ s \\
Oldroyd-B & $4{:}1$ contraction & 80 & 64 & $Re\in[0.4,80]$, $Wi=0.01\,Re$ & $U_{in}=2.5\times10^{-4}Re~\mathrm{m/s}$ & 51 & $t\in[0,5]$ s \\
Bingham & $4{:}1$ contraction & 150 & 134 & $Re\in[0.4,60]$, $Bi\in[0,5]$ & $U_{in}=2.5\times10^{-4}Re~\mathrm{m/s}$ & 51 & $t\in[0,5]$ s \\
\hline
Newtonian (source) & Triangular obstacle & 200 & --- & $Re\in[10,100]$ & $U=2\times10^{-4}Re~\mathrm{m/s}$ & 101 & $t\in[0,10]$ s \\
Oldroyd-B & Triangular obstacle & 200 & 168 & $Re\in[10,100]$, $Wi=0.01\,Re$ & $U=2\times10^{-4}Re~\mathrm{m/s}$ & 101 & $t\in[0,10]$ s \\
\hline
\end{tabular}
\end{table*}

In both fields, the errors of the from-scratch model are concentrated at the re-entrant corner and along the downstream wall, where the flow accelerates into the narrow channel.
For the velocity magnitude, these errors reach $18$--$39\%$ of the maximum velocity, compared with $9$--$27\%$ for the fine-tuned predictions.
The shear stress separates the two models more strongly than the velocity, with relative errors that are reduced by pretraining by a factor of $2.4$--$3$ (Table~\ref{tab:si_fields}).
The Giesekus case shows the clearest constitutive effect: shear thinning lowers the downstream shear stress by about an order of magnitude relative to the Newtonian flow at the same $Re$, and the fine-tuned model reproduces this reduced stress level and the low-stress layer near the centerline, whereas the from-scratch prediction does not.
Away from the corner, the fine-tuned predictions contain small-amplitude noise in the low-stress regions upstream of the contraction, which is visible on the logarithmic color scale but contributes little to the relative error.

\begin{figure*}[b]
    \centering
    \includegraphics[width=\textwidth]{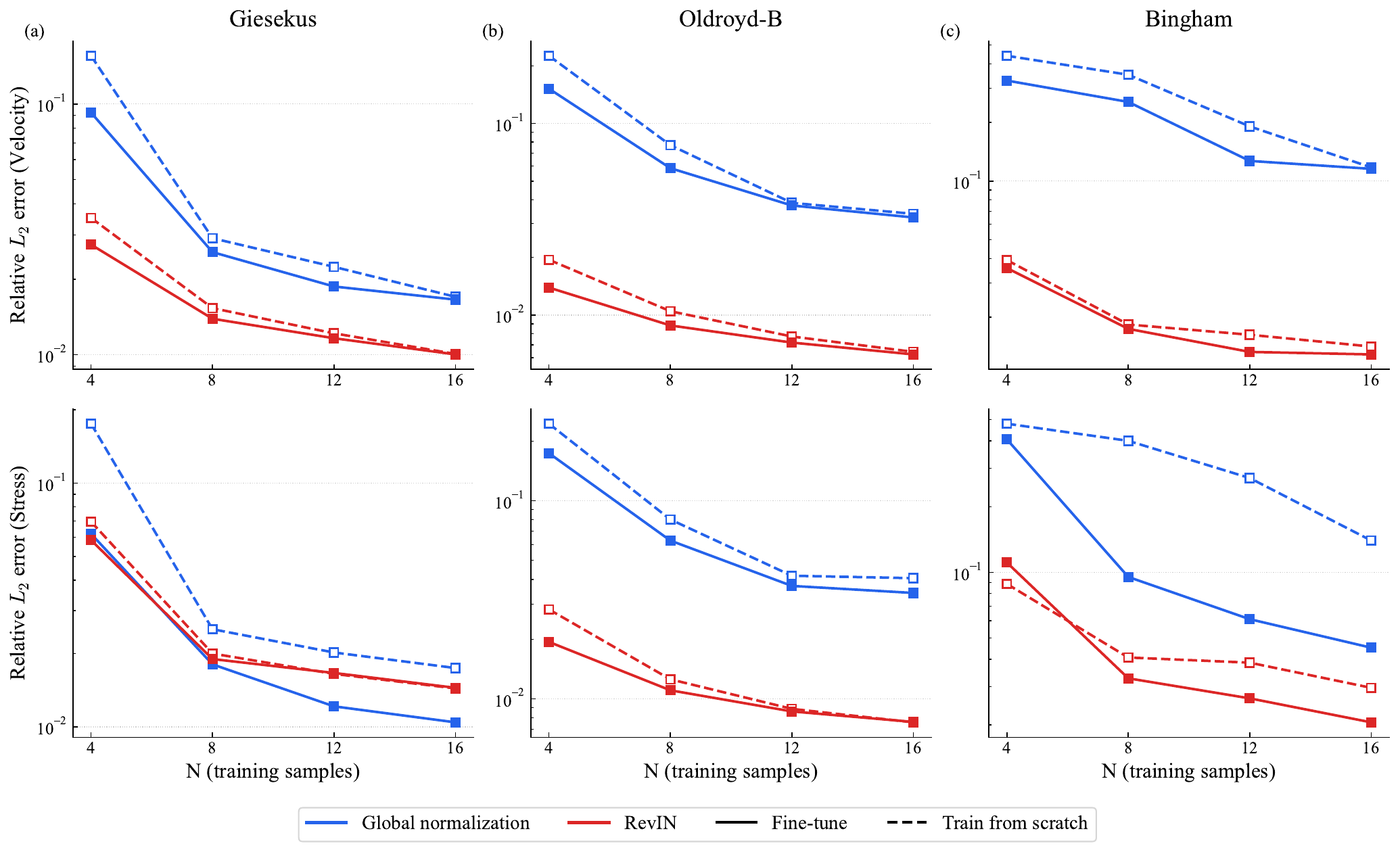}
    \caption{\textbf{Contraction-flow prediction with a ten-snapshot input window.} Real-space relative $L_2$ velocity (top row) and stress (bottom row) error as a function of the number of fine-tuning samples, $N$, for Giesekus, Oldroyd-B, and Bingham fluids. Global normalization and RevIN are distinguished by color, while pretrained and from-scratch models are shown by solid and dashed curves, respectively. Values are averaged over three seeds.}
    \label{fig:si_longwindow}
\end{figure*}

\begin{figure*}[b]
    \centering
    \includegraphics[width=\textwidth]{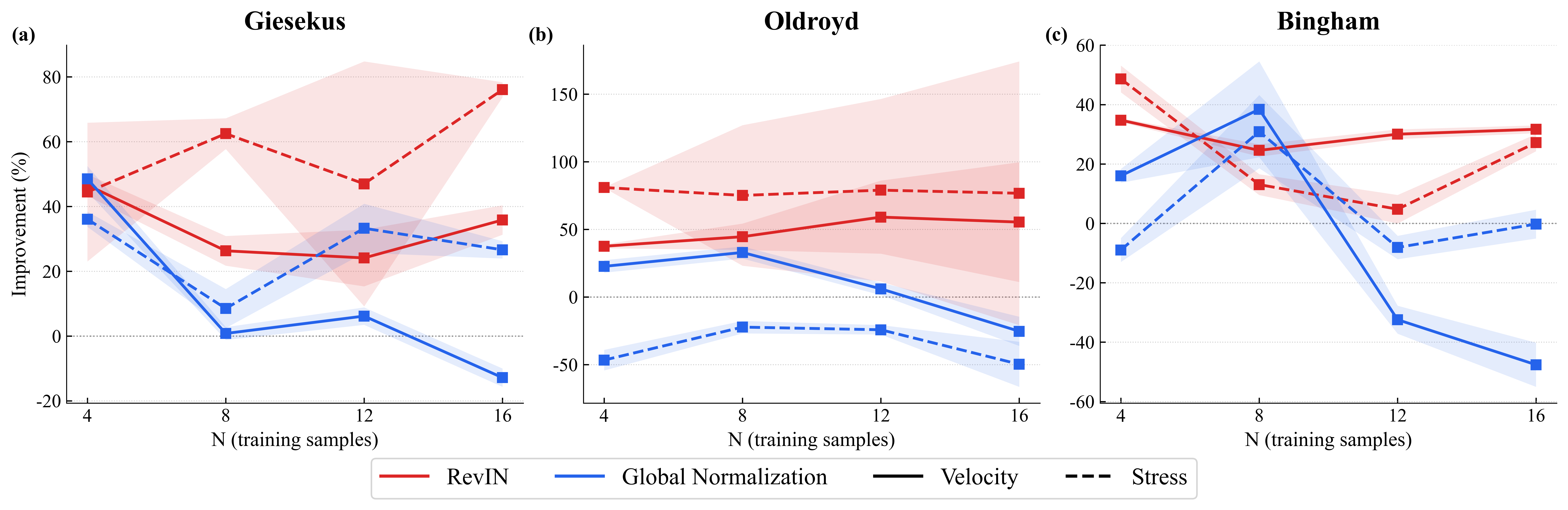}
    \caption{\textbf{Improvement from Newtonian pretraining in the contraction flows.} Percentage reduction in the real-space relative $L_2$ error obtained by Newtonian pretraining relative to training from scratch, $100\,(L_{\mathrm{scratch}}-L_{\mathrm{pretrained}})/L_{\mathrm{scratch}}$, as a function of the number of fine-tuning samples, $N$, for (a) Giesekus, (b) Oldroyd-B, and (c) Bingham fluids in the $4{:}1$ contraction with the two-snapshot input. Red and blue curves correspond to RevIN and Global normalization, and solid and dashed curves to the velocity and stress errors, respectively. Positive values indicate that pretraining reduces the error, and the dotted line marks zero improvement. Markers show the improvement computed from the seed-averaged errors, and shaded bands show the corresponding uncertainty propagated from the seed-to-seed standard deviations of the fine-tuned and from-scratch errors. The wide bands for Oldroyd-B under RevIN reflect single failed from-scratch runs at $N\ge8$ (main text, Sec.~III~B).}
    \label{fig:si_longwindow}
\end{figure*}

\begin{figure*}[b]
    \centering
    \includegraphics[width=\textwidth]{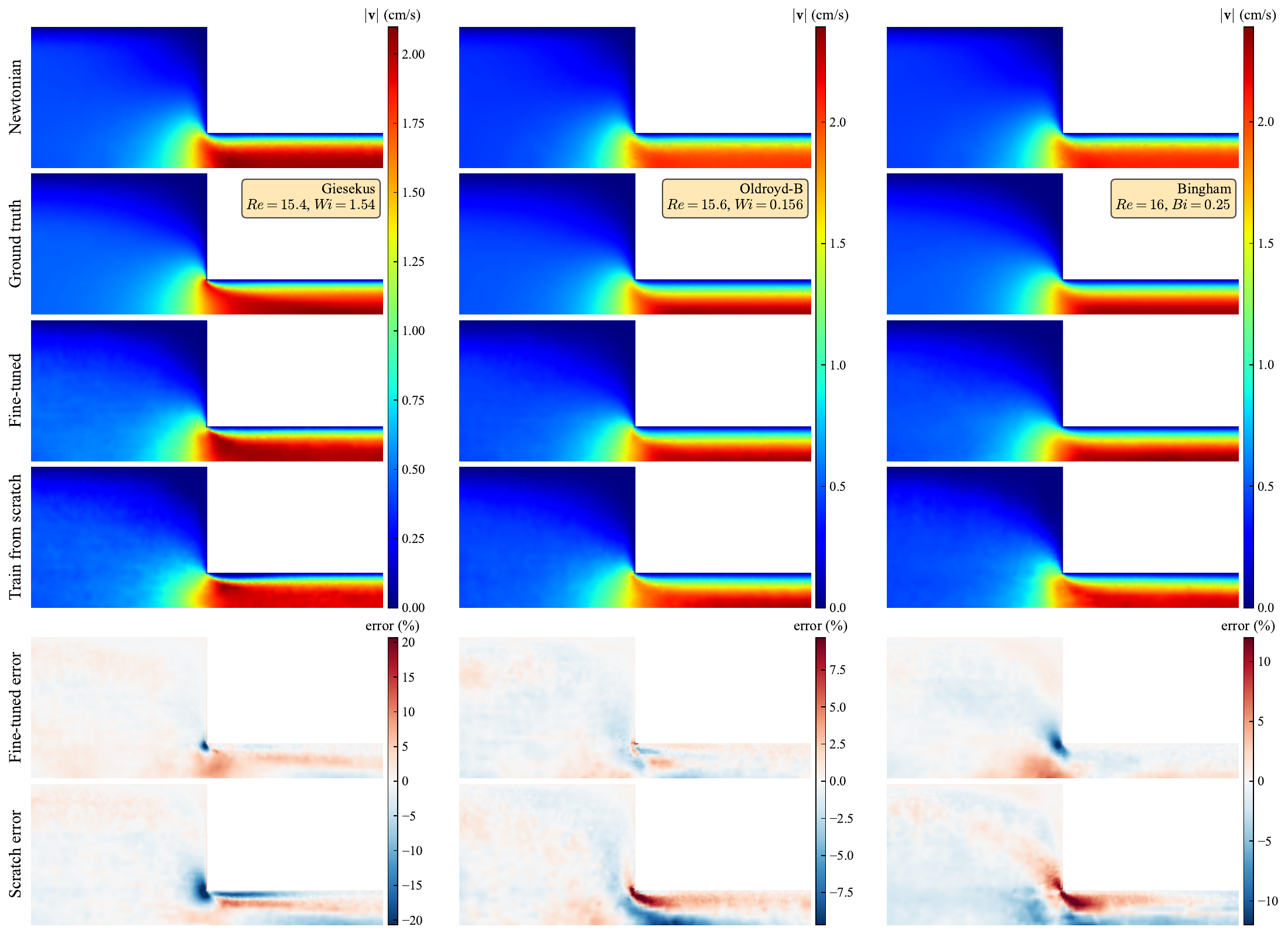}
    \caption{\textbf{Velocity fields and normalized error maps.} Velocity magnitude, $|\mathbf{v}|$, for the representative cases of Fig.~2 of the main text: Giesekus ($Re=15.4$, $Wi=1.54$), Oldroyd-B ($Re=15.6$, $Wi=0.156$), and Bingham ($Re=16$, $Bi=0.25$). Rows 1--4 show the Newtonian pretraining simulation at the same $Re$, the ground truth, the fine-tuned prediction, and the from-scratch prediction on a shared color scale. Rows 5 and 6 show the normalized errors of the fine-tuned and from-scratch predictions, $(|\mathbf{v}|-|\mathbf{v}|_{\mathrm{GT}})/\max|\mathbf{v}|_{\mathrm{GT}}$, in percent.}
    \label{fig:si_vmag}
\end{figure*}

\begin{figure*}[b]
    \centering
    \includegraphics[width=\textwidth]{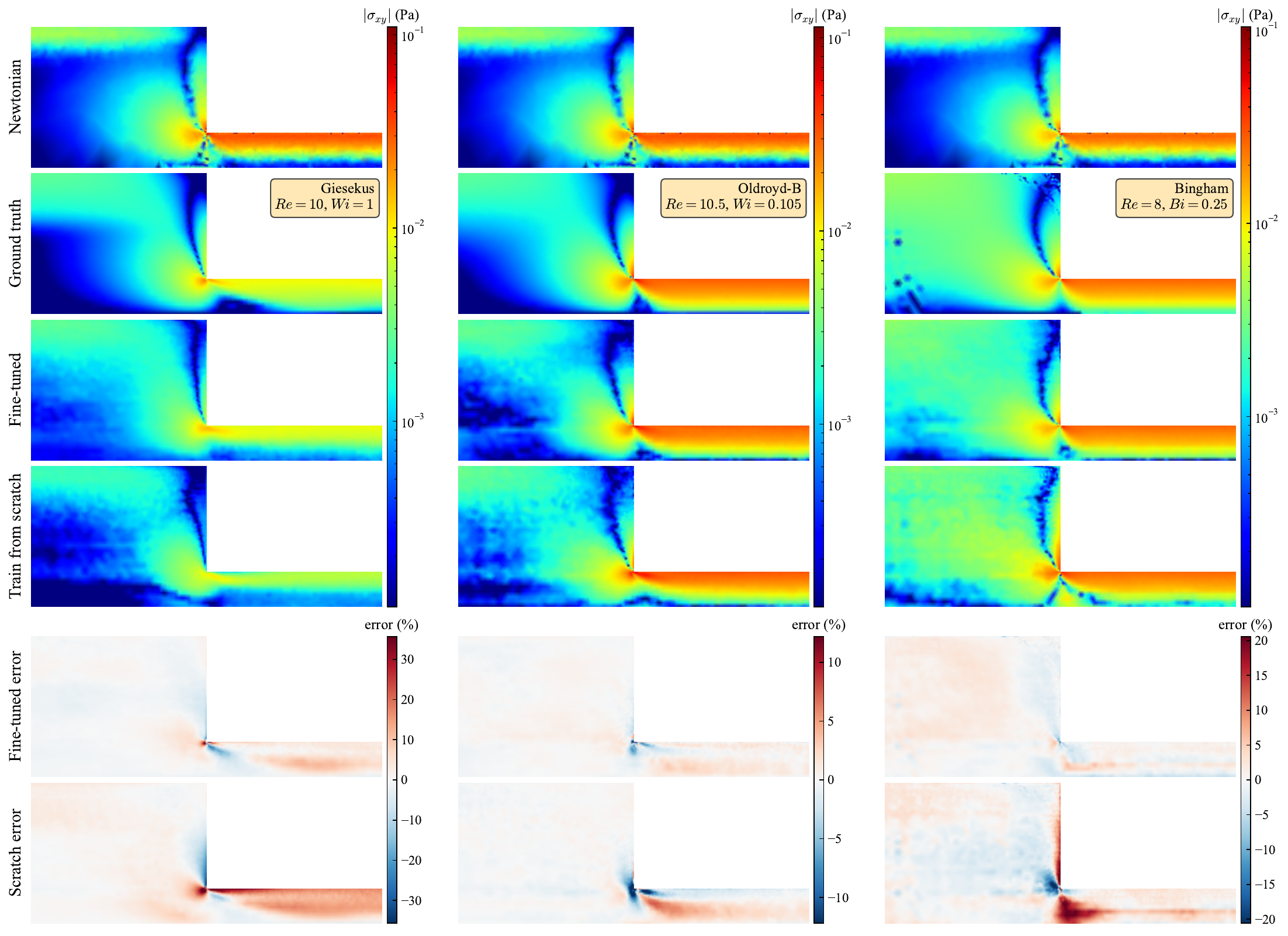}
    \caption{\textbf{Shear-stress fields and normalized error maps.} Magnitude of the shear stress, $|\sigma_{xy}|$, on a logarithmic color scale for Giesekus ($Re=10$, $Wi=1$), Oldroyd-B ($Re=10.5$, $Wi=0.105$), and Bingham ($Re=8$, $Bi=0.25$) fluids. Rows 1--4 show the Newtonian pretraining simulation at the same $Re$, the ground truth, the fine-tuned prediction, and the from-scratch prediction on a shared color scale, with a lower limit of $10^{-3}$ times the column maximum. Rows 5 and 6 show the normalized errors of the signed shear stress, $(\sigma_{xy}-\sigma_{xy,\mathrm{GT}})/\max|\sigma_{xy,\mathrm{GT}}|$, in percent.}
    \label{fig:si_sxy}
\end{figure*}

\begin{table*}
\centering
\caption{Target cases used for fine-tuning at each budget $N$. Contraction cases are listed by $Re$ (with $Wi=0.1\,Re$ for Giesekus and $Wi=0.01\,Re$ for Oldroyd-B) or by $(Re,Bi)$ for Bingham, using the downstream convention of Sec.~\ref{si:constitutive}; triangular-obstacle cases are listed by $Wi$ ($Re=100\,Wi$). The sets for different $N$ are not nested in general.}
\label{tab:si_trainsets}
\footnotesize
\begin{tabular}{p{2.2cm}p{0.5cm}p{14.6cm}}
\hline
Dataset & $N$ & Fine-tuning cases \\
\hline
Giesekus & 4 & $Re=$ 1, 14, 33.8, 42 \\
 & 8 & $Re=$ 1, 4, 12, 16, 24, 33.8, 39, 42 \\
 & 12 & $Re=$ 1, 2, 8, 12, 16, 18, 24, 28, 31, 36, 39, 42 \\
 & 16 & $Re=$ 1, 2, 4, 8, 12, 14, 16, 18, 24, 28, 31, 33.8, 36, 39, 41, 42 \\
\hline
Oldroyd-B & 4 & $Re=$ 0.4, 23.16, 48.4, 80 \\
 & 8 & $Re=$ 0.4, 13.04, 18.52, 26.92, 42.08, 54.72, 69.88, 80 \\
 & 12 & $Re=$ 0.4, 6.72, 13.04, 18.52, 23.16, 26.92, 37.04, 42.08, 54.72, 62.32, 76.2, 80 \\
 & 16 & $Re=$ 0.4, 6.72, 13.04, 15.92, 18.52, 23.16, 26.92, 33.24, 37.04, 42.08, 48.4, 54.72, 62.32, 69.88, 76.2, 80 \\
\hline
Bingham & 4 & (0.4, 5), (8, 0), (8, 2.5), (40, 0.0625) \\
 & 8 & (0.4, 0), (0.4, 5), (3, 0.125), (3, 3), (28, 0.025), (28, 5), (40, 0.0625), (60, 1.5) \\
 & 12 & (0.4, 0), (0.4, 5), (1, 0.025), (1, 3.75), (3, 0.125), (3, 3), (8, 0), (8, 2.5), (28, 0.025), (28, 5), (40, 0.0625), (60, 1.5) \\
 & 16 & (0.4, 0), (0.4, 5), (0.6, 0.0125), (1, 0.025), (1, 3.75), (2, 0.0625), (3, 0.125), (3, 3), (4, 0.25), (8, 0), (8, 2.5), (16, 0.0125), (28, 0.025), (28, 5), (40, 0.0625), (60, 1.5) \\
\hline
Oldroyd-B & 4 & $Wi=$ 0.100, 0.394, 0.679, 0.999 \\
(triangle) & 8 & $Wi=$ 0.100, 0.276, 0.394, 0.512, 0.629, 0.747, 0.873, 0.999 \\
 & 12 & $Wi=$ 0.100, 0.159, 0.276, 0.335, 0.453, 0.552, 0.629, 0.679, 0.747, 0.873, 0.955, 0.999 \\
 & 16 & $Wi=$ 0.100, 0.159, 0.218, 0.276, 0.349, 0.394, 0.453, 0.552, 0.607, 0.629, 0.679, 0.747, 0.842, 0.873, 0.955, 0.999 \\
 & 24 & $Wi=$ 0.100, 0.127, 0.159, 0.218, 0.245, 0.276, 0.335, 0.362, 0.394, 0.453, 0.484, 0.512, 0.570, 0.602, 0.629, 0.679, 0.715, 0.747, 0.806, 0.833, 0.864, 0.919, 0.946, 0.999 \\
 & 32 & $Wi=$ 0.100, 0.127, 0.159, 0.181, 0.218, 0.245, 0.276, 0.304, 0.335, 0.362, 0.394, 0.421, 0.453, 0.484, 0.512, 0.543, 0.570, 0.602, 0.629, 0.656, 0.679, 0.715, 0.747, 0.774, 0.806, 0.833, 0.864, 0.887, 0.919, 0.946, 0.987, 0.999 \\
\hline
\end{tabular}
\end{table*}

\begin{table}[t]
\centering
\caption{Architectural and training hyperparameters.}
\label{tab:si_hyper}
\small
\begin{tabular}{lc}
\hline
\multicolumn{2}{l}{\textit{Architecture}} \\
Input embedding dimension & 32 \\
Encoder self-attention blocks & 2 \\
Attention heads & 2 \\
Latent channels & 32 \\
Decoder cross-attention blocks & 1 \\
Output embedding dimension & 32 \\
Positional-encoding scale & 8 \\
Dropout & 0.05 \\
Trainable parameters & $\approx6.8\times10^{6}$ \\
\hline
\multicolumn{2}{l}{\textit{Newtonian pretraining}} \\
Epochs (contraction / triangle) & 600 / 2000 \\
Batch size & 16 \\
Optimizer (weight decay) & AdamW ($10^{-4}$) \\
Schedule (max.\ learning rate) & one-cycle ($10^{-4}$) \\
\hline
\multicolumn{2}{l}{\textit{Fine-tuning and training from scratch}} \\
Epochs (contraction / triangle) & 500 / 2000 \\
Batch size & 4 \\
Optimizer & AdamW \\
Weight decay (fine-tuned/scratch) & $10^{-6}$ / $10^{-4}$ \\
Schedule (max.\ learning rate) & one-cycle ($10^{-4}$) \\
Warm-up fraction & 0.3 \\
Initial / final division factor & $10^{4}$ / $10^{6}$ \\
Checkpoint, fine-tuned & best validation \\
Checkpoint, from scratch & final epoch \\
Seeds & 3 \\
\hline
\end{tabular}
\end{table}

\begin{table}[b]
\centering
\caption{Relative $L_2$ errors of the representative cases in Figs.~\ref{fig:si_vmag} and \ref{fig:si_sxy}, evaluated over $|x|\le5$ at $t=5~\mathrm{s}$.}
\label{tab:si_fields}
\small
\begin{tabular}{llcc}
\hline
Field & Case & Fine-tuned & From scratch \\
\hline
$|\mathbf{v}|$ & Giesekus, $Re=15.4$ & 6.3\% & 12.8\% \\
 & Oldroyd-B, $Re=15.6$ & 2.7\% & 6.5\% \\
 & Bingham, $Re=16$, $Bi=0.25$ & 4.9\% & 7.9\% \\
\hline
$\sigma_{xy}$ & Giesekus, $Re=10$ & 18.8\% & 47.9\% \\
 & Oldroyd-B, $Re=10.5$ & 8.1\% & 19.4\% \\
 & Bingham, $Re=8$, $Bi=0.25$ & 6.4\% & 18.9\% \\
\hline
\end{tabular}
\end{table}

\begin{table*}[b]
\centering
\caption{Real-space relative $L_2$ errors and improvement from Newtonian pretraining with a ten-snapshot input window, averaged over three seeds.}
\label{tab:si_longwindow}
\begin{tabular}{llc|cc|cc|cc}
\hline
 & & & \multicolumn{2}{c|}{Fine-tuned} & \multicolumn{2}{c|}{From scratch} & \multicolumn{2}{c}{Improvement (\%)} \\
Fluid & Normalization & $N$ & Velocity & Stress & Velocity & Stress & Velocity & Stress \\
\hline
Giesekus & Global & 4 & 0.092 & 0.062 & 0.156 & 0.175 & 40.7 & 64.7 \\
 & & 8 & 0.026 & 0.018 & 0.029 & 0.025 & 12.0 & 28.2 \\
 & & 12 & 0.019 & 0.012 & 0.022 & 0.020 & 16.5 & 39.9 \\
 & & 16 & 0.017 & 0.010 & 0.017 & 0.017 & 2.5 & 40.1 \\
 & RevIN & 4 & 0.028 & 0.058 & 0.035 & 0.070 & 21.8 & 16.0 \\
 & & 8 & 0.014 & 0.019 & 0.015 & 0.020 & 9.3 & 5.1 \\
 & & 12 & 0.012 & 0.017 & 0.012 & 0.017 & 4.4 & $-0.9$ \\
 & & 16 & 0.010 & 0.015 & 0.010 & 0.014 & 0.4 & $-0.6$ \\
\hline
Oldroyd-B & Global & 4 & 0.152 & 0.174 & 0.226 & 0.245 & 32.7 & 29.0 \\
 & & 8 & 0.058 & 0.063 & 0.077 & 0.080 & 24.1 & 21.7 \\
 & & 12 & 0.037 & 0.037 & 0.039 & 0.042 & 3.1 & 11.1 \\
 & & 16 & 0.032 & 0.034 & 0.034 & 0.041 & 4.2 & 15.8 \\
 & RevIN & 4 & 0.014 & 0.019 & 0.020 & 0.028 & 28.4 & 31.4 \\
 & & 8 & 0.009 & 0.011 & 0.011 & 0.013 & 15.7 & 12.0 \\
 & & 12 & 0.007 & 0.009 & 0.008 & 0.009 & 7.0 & 2.9 \\
 & & 16 & 0.006 & 0.008 & 0.006 & 0.008 & 3.1 & $-0.5$ \\
\hline
Bingham & Global & 4 & 0.328 & 0.407 & 0.440 & 0.479 & 25.4 & 15.2 \\
 & & 8 & 0.255 & 0.095 & 0.352 & 0.400 & 27.5 & 76.2 \\
 & & 12 & 0.127 & 0.061 & 0.191 & 0.270 & 33.5 & 77.4 \\
 & & 16 & 0.116 & 0.045 & 0.118 & 0.140 & 1.7 & 67.7 \\
 & RevIN & 4 & 0.036 & 0.111 & 0.039 & 0.089 & 9.0 & $-25.3$ \\
 & & 8 & 0.017 & 0.033 & 0.018 & 0.041 & 5.0 & 19.7 \\
 & & 12 & 0.013 & 0.027 & 0.016 & 0.039 & 18.4 & 31.4 \\
 & & 16 & 0.013 & 0.021 & 0.014 & 0.030 & 8.9 & 30.5 \\
\hline
\end{tabular}
\end{table*}